\documentclass[acmtog]{acmart}
\acmSubmissionID{221s2}

\usepackage{soul}
\usepackage{booktabs} 

\usepackage[ruled]{algorithm2e} 

\SetAlFnt{\small}
\SetAlCapFnt{\small}
\SetAlCapNameFnt{\small}
\SetAlCapHSkip{0pt}

\acmJournal{TOG}
\acmVolume{39}
\acmNumber{6}
\acmArticle{263} 
\acmYear{2020}
\acmMonth{12}

\acmDOI{10.1145/3414685.3417806}

\newcommand*{\ShowNotes}{}
\ifdefined\ShowNotes
  \newcommand{\colornote}[3]{{\color{#1}{#2 #3}\normalfont}}
\else
  \newcommand{\colornote}[3]{}
\fi

\begin{document}
\title{MeshWalker: Deep Mesh Understanding by Random Walks}

\author{Alon Lahav}
\affiliation{%
  \institution{Technion – Israel Institute of Technology}
}
\email{alon.lahav2@gmail.com}
\author{Ayellet Tal}
\affiliation{%
  \institution{Technion – Israel Institute of Technology}
}
\email{ayellet@ee.technion.ac.il}


\begin{abstract}
Most attempts to represent  3D shapes for deep learning have focused on volumetric grids, multi-view images and point clouds.
In this paper we look at the most popular representation of 3D shapes in computer graphics---a triangular mesh---and ask how it can be utilized within deep learning.
The few attempts to answer this question propose to adapt convolutions \& pooling to suit {\em Convolutional Neural Networks (CNNs)}.
This paper proposes a very different approach, termed {\em MeshWalker} 
to learn the shape directly from a given mesh.
The key idea is to represent the mesh by random walks along the surface, which "explore" the mesh's geometry and topology.
Each walk is organized as a list of vertices, which in some manner imposes regularity on the mesh.
The walk is fed into  a {\em Recurrent Neural Network (RNN)} that "remembers" the history of the walk.
We show that our approach achieves state-of-the-art results for two fundamental shape analysis tasks: shape classification and semantic segmentation.
Furthermore, even a very small number of examples suffices for learning.
This is highly important, since large datasets of meshes are difficult to acquire.
\end{abstract}


\begin{teaserfigure}
\centering  
\fbox{\includegraphics[width=0.97\textwidth, trim={0cm 4.0cm 0cm 3.0cm}, clip]{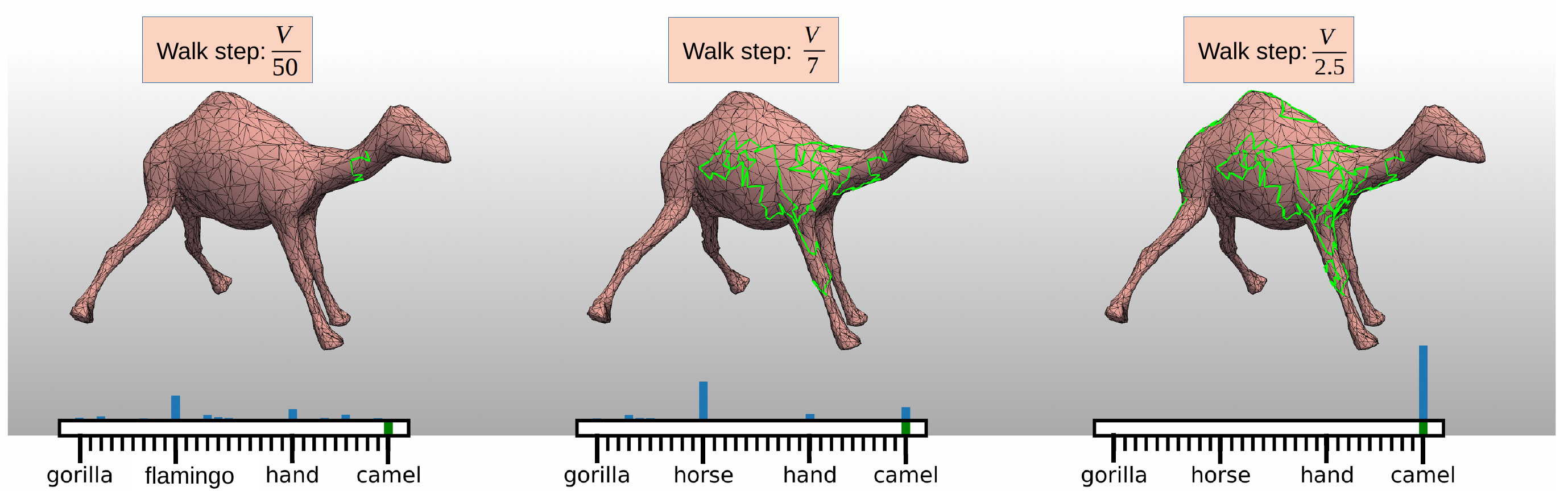}}
  \caption{{\bf Classification by MeshWalker.}
    This figure shows classification results as the walk (in green) proceeds along the surface of a camel ($4K$ faces) from SHREC11~\cite{lian2011shape}.
    The initial point was randomly chosen on the neck.
    After $V/50$ steps (left), $V$ being the number of vertices, the system is uncertain regarding the class, and the highest probability predictions are for the flamingo class and for the hand class (out of $30$ classes).
    After continuing the random walk along the body and the front leg for $V/7$ steps, the probability of being a horse is higher than before, but the camel already has quite a high probability.
    Finally, after $V/2.5$ steps (right) and walking also along the hump, the system correctly classifies the model as a camel.
     }
\label{fig:teaser}
\end{teaserfigure}

\maketitle

\section{Introduction}
\label{sec:introduction}

The most-commonly used representation of surfaces in computer graphics is a polygonal mesh, due to its numerous benefits, including efficiency and high-quality.
Nevertheless, in the era of deep learning, this representation is often bypassed because of its irregularity, which does not suit {\em Convolutional Neural Networks (CNNs)}.
Instead, 3D data is often represented as volumetric grids~\cite{maturana2015voxnet, DBLP:journals/corr/AlvarZB16, roynard2018classification, ben20183dmfv} or multiple 2D projections~\cite{su2015multi, boulch2017unstructured, 
Feng_2018_CVPR, DBLP:journals/corr/abs-1811-01571, kanezaki2018rotationnet}.
In some recent works point clouds are utilized and new ways to convolve or pool are proposed~\cite{atzmon2018point, xu2018spidercnn, li2018pointcnn, hua2018pointwise, thomas2019kpconv}.

Despite the benefits of these representations, they miss the notions of neighborhoods and connectivity and might not be as good for capturing local surface properties.
Recently, several works have proposed to maintain the potential of the mesh representation, while still utilizing neural networks.
FeaStNet~\cite{verma2018feastnet} proposes a graph neural network in which the neighborhood of each vertex for the convolution operation is calculated dynamically based on its features.
MeshCNN~\cite{hanocka2019meshcnn} defines pooling and convolution layers over the mesh edges.
MeshNet~\cite{feng2019meshnet} treats the faces of a mesh as the basic unit and extracts their spatial and structural features individually to offer the ﬁnal semantic representation.
LRF-Conv~\cite{yang2020continuous} learns descriptors directly from the raw mesh by defining new continuous convolution kernels that provide robustness to sampling.
All these methods redefine the convolution operation, and by doing so, are able to fit the unordered structure of a mesh to a CNN framework.

We propose a novel and fundamentally different approach,  named {\em MeshWalker}.
As in previous approaches that learn directly from the mesh data, the basic question is how to impose regularity on the unordered data.
Our key idea is to represent the mesh by random walks on its surface.
These walks explore the local geometry of the surface, as well as its global one.
Every walk is fed into a {\em Recurrent Neural Network (RNN)}, that "remembers" the walk's history.

In addition to simplicity, our approach has three important benefits. 
First, we will show that even a small dataset suffices for training.
Intuitively, we can generate multiple random walks for a single model; these walks provide multiple explorations of the model. 
This may be considered as equivalent to using different projections of 3D objects in the case of image datasets.
Second, as opposed to CNNs, RNNs are inherently robust to sequence length.
This is vital in the case of meshes, as datasets include objects of various granularities.
Third, the meshes need not be watertight or have a single connected component; our approach can handle any triangular mesh.

Our approach is general and can be utilized to address a variety of shape analysis tasks.
We demonstrate its benefit in two basic applications: mesh classification and mesh semantic segmentation.
Our results are superior to those of state-of-the-art approaches on common datasets and on highly non-uniform meshes.
Furthermore, when the training set is limited in size, the accuracy improvement over the state-of-the-art methods is highly evident.

Hence, this paper makes three contributions:
\begin{enumerate}
    \item 
    We propose a novel representation of meshes for neural networks: random walks on surfaces.
    \item
    We present an end-to-end learning framework that realizes this representation within RNNs.
    We show that this framework works well even when the dataset is very small.
    This is important in the case of 3D, where large datasets are seldom available and are difficult to generate.
    \item 
    We demonstrate the benefits of our method in two key applications: 3D shape classification and semantic segmentation.
\end{enumerate}

\section{Related Work}
\label{sec:relatedwork}
Our work is at the crossroads of three fields, as discussed below.

\subsection{Representing 3D objects for Deep Neural Networks}
\label{subsec:related_work_3d_representations}
A variety of representations of 3D shapes have been proposed in the context of deep learning.
The main challenge is how to re-organize the shape description such that it could be processed within deep learning frameworks. 
Hereafter we briefly review the main representations;
see~\cite{gezawa2020review} for a recent excellent survey.
 
\paragraph{Multi-view 2D projections.}
This representation is essentially a set of 2D images, each of which is a rendering of the object from a different viewpoint~\cite{su2015multi, kalogerakis20173d, qi2016volumetric, sarkar2018learning, gomez2017lonchanet, johns2016pairwise, zanuttigh2017deep, bai2016gift, wang2019dominant, kanezaki2018rotationnet, feng2018gvcnn, he2018triplet, han20193d2seqviews}.
The major benefit of this representation is that it can naturally utilize any image-based CNN. 
In addition, high-resolution inputs can be easily handled. 
However,  it is not easy to determine the optimal number of views; if that number is large, the computation might be costly.
Furthermore, self-occlusions might be a drawback.

\paragraph{Volumetric grids.}
These grids are analogous to the 2D grids of images.
Therefore, the main benefit of this representation is that operations that are applied on 2D grids can be extended to 3D  in a straightforward manner~\cite{wu20153d, brock2016generative, tchapmi2017segcloud, fanelli2011real, maturana2015voxnet, wang2019normalnet, sedaghat2016orientation, zhi2018toward}.
The primary drawbacks of volumetric grids are their limited resolution and the heavy computation cost needed.

\paragraph{Point clouds.}
This representation consists of a set of 3D points, sampled from the object's surface. 
The simplicity, close relationship to data acquisition, and the ease of conversion from other representations, make  point clouds an attractive representation.
Therefore, a variety of recent works proposed successful techniques for point cloud shape analysis using neural networks~\cite{qi2017pointnet, qi2017pointnet++, wang2019dynamic, guerrero2018pcpnet, williams2019deep, atzmon2018point, li2018pointcnn, liu2019relation, xu2019geometry, zhu2019random}.
These methods attempt to learn a representation for each point, using its neighbors (Euclidean-wise) either by multi layer perceptions or by convolutional layers.
Some also  define novel pooling layers.
Point cloud representations might fall short in applications when the connectivity is highly meaningful (e.g. segmentation) or when the salient information is concentrated in small specific areas.

\paragraph{Triangular meshes.}
This representation is the most widespread representation in computer graphics and the focus of our paper.
The major challenge of using meshes within deep learning frameworks is the irregularity of the representation---each vertex has a different number of neighbors, at different distances.

The pioneering work of~\cite{masci2015geodesic} introduces deep learning of local features and shows
 how to make the convolution operations intrinsic to the mesh.
In~\cite{poulenard2018multi} a new convolutional layer is defined, which allows the propagation of geodesic information throughout the network layers.
FeaStNet~\cite{verma2018feastnet} proposes a graph neural network in which the neighborhood of each vertex for the convolution operation is calculated dynamically based on its features.
Another line of works exploits the fact that local patches are approximately Euclidean. 
The 3D  manifolds are then parameterized  in 2D, where standard CNNs are utilized~\cite{henaff2015deep, sinha2016deep, boscaini2016learning, maron2017convolutional, ezuz2017gwcnn, haim2019surface}.
A different approach is to apply a linear map to a spiral of neighbors~\cite{gong2019spiralnet++, lim2018simple}, which works well for meshes with a similar graph structure.

\begin{figure*}[tb]
\centering 
\begin{tabular}{ccc}
\includegraphics[height=0.33\textwidth]{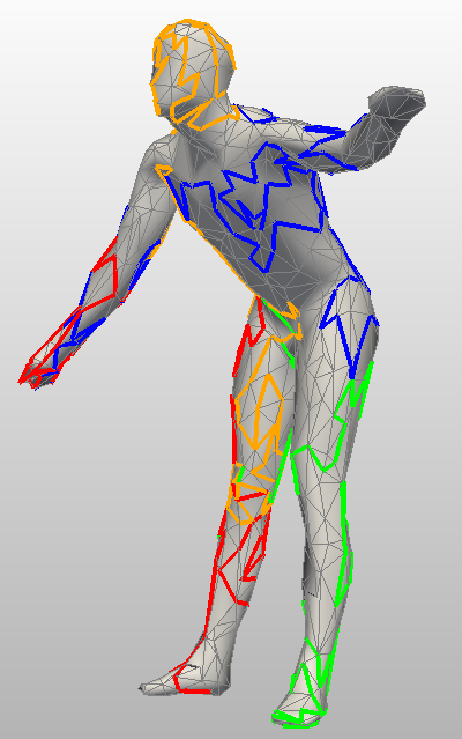} &
\includegraphics[height=0.33\textwidth]{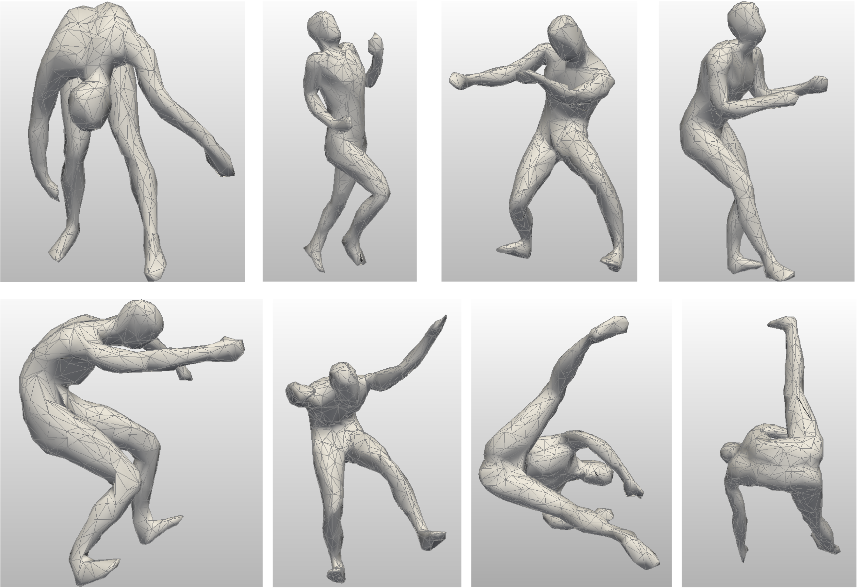} &
\includegraphics[height=0.33\textwidth]{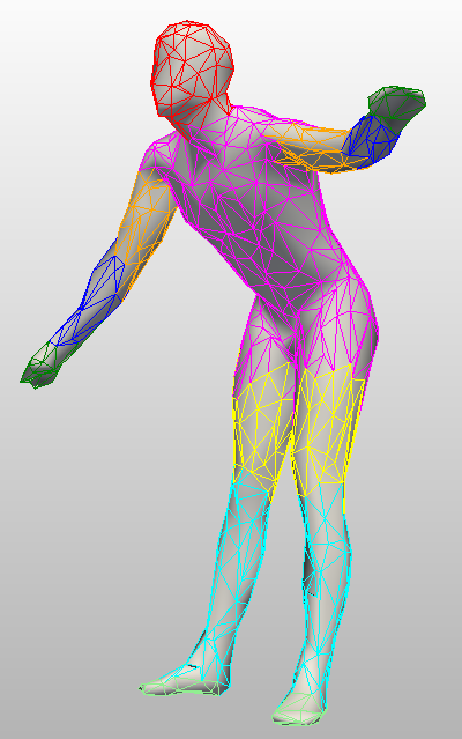}  \\
(a) $5$ walks on the surface & (b) Classification: Samples from the class the input belongs to& (c) Semantic segmentation
\end{tabular}
\caption{{\bf Outline.}
To explore a mesh, walks on its surface are generated and study the surface both locally and globally (a).
These walks provide sufficient information to perform shape analysis tasks, such as classification and segmentation.
Specifically,~(b) shows samples from the class to which MeshWalker correctly classified the model from~(a) and~(c) shows the resulting segmentation.
The models are from SHREC11~\cite{lian2011shape}.
}
\label{fig:outline} 
\end{figure*}

Two approaches were recently introduced:
MeshNet~\cite{feng2019meshnet} treats faces of a mesh as the basic unit and  extracts their spatial and structural features individually, to offer the final semantic representation.
MeshCNN~\cite{hanocka2019meshcnn} is based on a very unique idea of using the edges of the mesh to perform pooling and convolution. 
The convolution operations exploit the regularity of edges---having $4$ edges of their incidental triangles.
An edge collapse operation is used for pooling, which maintains surface topology and generates new mesh connectivity for further convolutions.

\subsection{Classification}
\label{subsec:related_work_classification}
Object classification refers to the task of classifying a given shape into one of pre-defined categories.
Before deep learning methods became widespread, the main challenges were finding good descriptors and good distance functions between these descriptors.
According to the thorough  review of~\cite{lian2013comparison}, the methods could be roughly classified into algorithms employing local features~\cite{johnson1999using, lowe2004distinctive, liu2006shape, sun2009concise, ovsjanikov2009shape}, topological structures~\cite{hilaga2001topology, sundar2003skeleton, tam2007deformable}, isometry-invariant global geometric properties~\cite{reuter2005laplace, jain2007spectral, mahmoudi2009three}, 
direct shape matching, or canonical forms~\cite{memoli2005theoretical, memoli2007use, bronstein2006efficient, elad2003bending}.

Many of the recent techniques already use deep learning for classification.
They are described in Section~\ref{subsec:related_work_3d_representations}, for instance~\cite{hanocka2018alignet, qi2017pointnet, qi2017pointnet++, li2018pointcnn, ezuz2017gwcnn, bronstein2011shape, feng2019meshnet, thomas2019kpconv, liu2019relation, velivckovic2017graph, wang2019graph, kipf2016semi, perozzi2014deepwalk}.

\subsection{Semantic segmentation}
\label{subsec:related_work_segmentatio}
Mesh segmentation is a key ingredient in many computer graphics tasks, including modeling, animation and a variety of shape analysis tasks.
The goal is to determine, for the basic elements of the mesh (vertex, edge or face), to which segment they belong.
Many approaches were proposed, including region growing~\cite{chazelle1997strategies, lavoue2005new, zhou2004decomposing, koschan2003perception, sun2002triangle, katz2005mesh},
clustering~\cite{shlafman2002metamorphosis, katz2003hierarchical, gelfand2004shape, attene2006hierarchical}, spectral analysis \cite{alpert1995spectral, gotsman2003graph, liu2004segmentation, zhang2005mesh} and more.
See~\cite{Attene_06,shamir2008survey,rodrigues2018part} for excellent surveys of segmentation methods.

Lately, deep learning has been utilized for this task as well.
Each proposed approach handles a specific shape representation, as described in Section~\ref{subsec:related_work_3d_representations}.
These approaches include among others ~\cite{hanocka2018alignet, qi2017pointnet, qi2017pointnet++, li2018pointcnn, yang2020continuous, haim2019surface, maron2017convolutional, qi2017pointnet++, guo20153d}.

\section{MeshWalker outline}
\label{sec:outline}

Imagine an ant walking on a surface; it will "climb" on ridges and go through valleys. 
Thus, it will explore the local geometry of the surface, as well as the global terrain. 
Random walks have been shown to incorporate both global and local information about a given object~\cite{Lai:2008:FMS:1364901.1364927,lovasz1993random, grady2006random, noh2004random}. 
This information may be invaluable for shape analysis tasks, nevertheless, random walks have not been used to represent meshes within a deep learning framework before. 

Given a polygonal mesh, we propose to randomly walk through the vertices of the mesh, along its edges,
as shown in Fig.~\ref{fig:outline}(a).
In our ant analogy, the longer the walk, the more information is acquired by the ant.
But how shall this information be accumulated?
We propose to feed this representation into a Recurrent Neural Network (RNN) framework, which aggregates properties of the walk.
This aggregated information will enable the ant to perceive the shape of the mesh.
This is particularly beneficial for shape analysis tasks that require both the 3D global structure and some local information of the mesh, as demonstrated in Fig.~\ref{fig:outline}(b-c).

Algorithm~\ref{alg:MeshWalkerTraining} describes the training procedure of our proposed {\em MeshWalker} approach. 
A defining property of it is that the \textbf{same} piece of algorithm is used for every vertex along the walk (i.e., each vertex the ant passes through).
The algorithm  iterates on the following:
A mesh is first extracted from the dataset (it could be a mesh that was previously extracted).
A vertex is chosen randomly as the head of the walk and then a random walk is generated.
This walk is the input to an RNN model.
Finally, the RNN model's parameters~$\theta$ are updated by minimizing the {\em Softmax} cross entropy loss $L$, using Adam optimizer~\cite{kingma2014adam}. 

\begin{algorithm}[t]
\SetAlgoNoLine
\KwIn{Labeled mesh dataset, $\mathcal{M}$ }
\KwOut{$\theta$---RNN model parameters}
$\theta_0 \gets RNN$ random parameters\;
$\mathcal{M} \gets$ MeshPreprocessing($\mathcal{M}$)\;
\Repeat{Convergence} 
    {
        $(M_i, y_i) \gets$ random mesh $M_i  \in \mathcal{M}$ and label(s) $y_i$\;
        $v_{ij} \gets$ random starting vertex\;
        ${w_{ij} \gets GenerateWalk(M_i, v_{ij})}$\;
        ${x}_{ij} \gets RepresentWalk(M_i, w_{ij})$\;
        $\theta_i \gets learningFromWalks(\theta_{i-1}, {x}_{ij}, y_i)$\;
    }
\caption{MeshWalker Training}
\label{alg:MeshWalkerTraining}
\end{algorithm}

Section~\ref{sec:model} elaborates on the architecture of our MeshWalker learning model, as well as on each of the  ingredients of the iterative step.
Section~\ref{subsec:preprocessing} explains the mesh pre-processing step, which essentially performs mesh simplification, and provides implementation details.

\section{Learning to walk over a surface}
\label{sec:model}
This section explains how to realize Algorithm~\ref{alg:MeshWalkerTraining}.
It begins by elaborating on the construction of a random walk on a mesh.
It then proceeds to describe the network that learns from walks in order to understand meshes.

\begin{figure*}[ht]
\centering 
\includegraphics[width=0.9\textwidth]{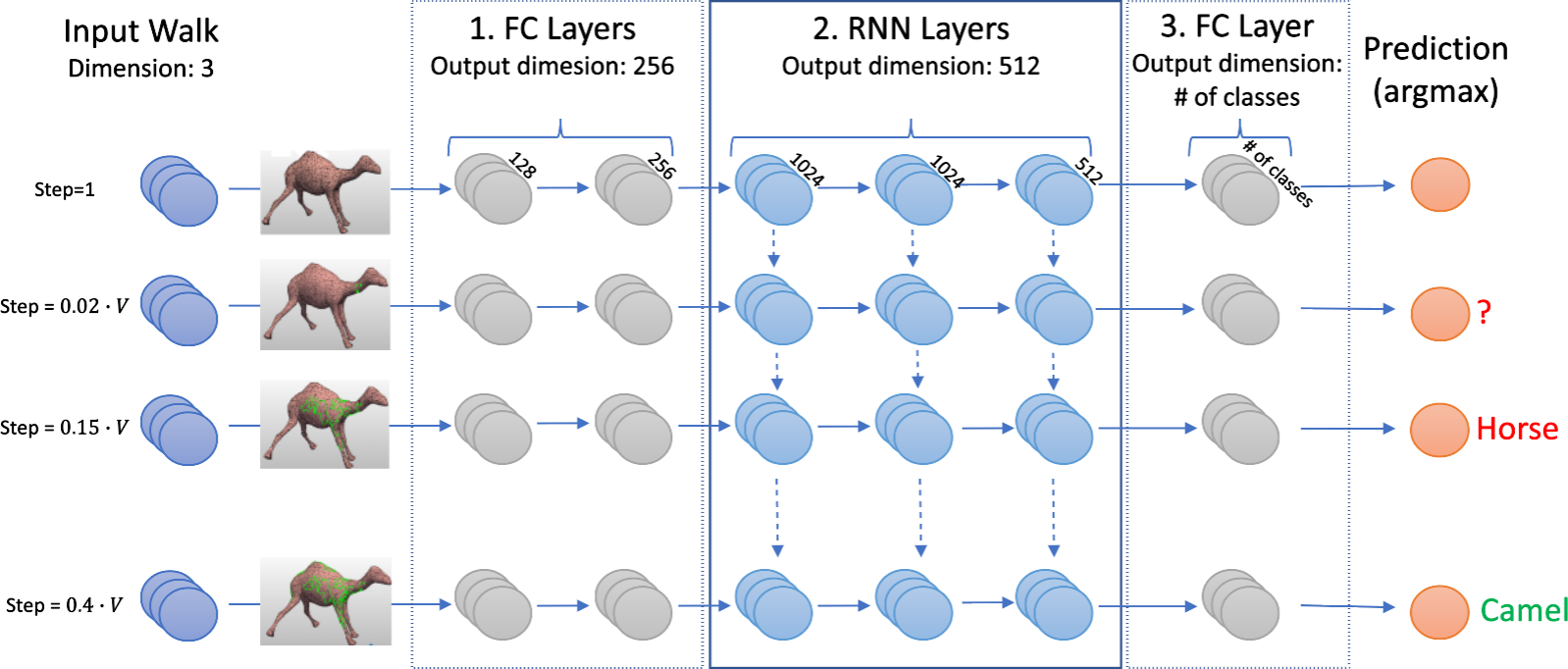} 
\caption{{\bf Network architecture.} 
The network consists of three components: 
The first component (FC layers) changes the feature space; the second component (RNN layers) aggregates the information along the walk; and the third component (an FC layer) predicts the outcome of the network. 
For classification, the prediction of the last vertex of the walk is considered and {\em Softmax} is applied to its resulting vector (the bottom-right orange circle, classified as a camel).
For segmentation (not shown in this figure), the network is similar.
However, {\em Softmax} is applied to each of the resulting vectors of the vertices (the orange circles in the right column); each vertex is classified into a segment.
}
 \label{fig:NetworkArch}
\end{figure*}

\subsection{What is a walk?}
Walks provide a novel way to organize the mesh data.
A {\em walk} is a sequence of vertices (not necessarily adjacent), each of which is associated with basic information.

\paragraph{Walk generation.}
We adopt a very simple strategy to generate walks, out of many possible ones.
Recall that we are given the first vertex $v_{ij}$ of a walk.
Then, to generate the walk $w_{ij}$, the other vertices are iteratively added, as follows.
Given the current vertex of the walk, the next vertex is chosen randomly from its adjacent vertices (those that belong to its one-ring neighbors).

If such a vertex does not exist (as all the neighbors already belong to the walk),  the walk is tracked backwards until an un-visited neighbor is found; this neighbor is added to the walk. 
In this case, the walk is not a linear sequence of vertices connected via edges, but rather a tree.
If the mesh consists of multiple connected component, it is possible that the walk reaches a dead-end.
In this case, a new random  un-visited vertex is chosen and the walk generation proceeds as before.
We note that in all cases, the input to the RNN is a sequence of vertices, arranged by their discovery order.
In practice,  the length of the walk is set by default to $\lceil V/2.5 \rceil$, where $V$ is number of vertices. 

\paragraph{Walk representation.}
 Once the walk $w_{ij}$ is determined, the representation $x_{ij}$ of this walk should be defined; this would be the input to the RNN. 
Each vertex is represented as the 3D translation from the previous vertex in the walk (${\Delta X, \Delta Y, \Delta Z}$).
This is inline with the deep learning philosophy, which prefers end-to-end learning instead of hand-crafted features that are separated from a classifier, 
We note that we also tried other representations, including vertex coordinates, normals, and curvatures, but the results did not improve.

\paragraph{Walks at inference time.}
At inference, several walks are being used for each mesh.
Each walk produces a vector of probabilities to belong to the different classes (in the case of classification).
These vectors are averaged to produce the final result.
To understand the importance of averaging, let us consider the walks on the camel in Fig.~\ref{fig:teaser}.
Since walks are generated randomly, we expect some of them to explore atypical parts of the model, such as the legs, which are similar to horse legs.
Other walks, however, are likely to explore unique parts, such as the hump or the head.
The average result will most likely be the camel, as will be shown in Section~\ref{sec:applications}.

\subsection{Learning from walks}
\label{subsec:learning}
Once walks are defined, the next challenge is 
to distillate the information accumulated along a walk into a single descriptor vector.
Hereafter we  discuss the network architecture and the training. 

\paragraph{Network architecture.}
The model consists of three sub-networks, as illustrated in  Fig.~\ref{fig:NetworkArch}.
The first sub-network is given the current vertex of the walk and learns a new feature space, i.e.  it transforms the 3D input feature space into a 256D feature space.
This is done by two fully connected (FC) layers, followed by an {\em instance normalization}~\cite{ulyanov2016instance} layer and  {\em ReLu} as nonlinear activation;
both empirically outperform other alternatives.

The second sub-network is the core of our approach.
It utilizes a recurrent neural network (RNN) whose defining property is being able to "remember" and accumulate knowledge.
Briefly, a recurrent neural network~\cite{graves2008novel,hochreiter1997long,cho2014learning} is a  connectionist model that contains a self-connected hidden layer. 
The benefit of self-connection is that the ‘memory’ of previous inputs remains in the network’s internal state, allowing it to make use of past context.
In our setting, the RNN gets as input a feature vector  (the result of the previous sub-network), learns the hidden states that describe the walk up to the current vertex, and outputs a state vector that contains the  information gathered along the walk. 

Another benefit of RNNs, which is crucial in our case, is not being confined to fixed-length inputs or outputs.
Thus, we can use the model to inference on a walk of a certain length, which may differ from walk lengths the model was trained on.

To implement the RNN part of our model, we use three {\em Gated Recurrent Unit (GRU)} layers of~\cite{cho2014learning}.
Briefly, the goal of an GRU layer is to accumulate only the important information from the input sequence and to forget the non-important information.

Formally, let $x_t$ be the input at time $t$ and $h_t$ be the hidden state at time $t$;
let the {\em reset gate} $r_t$ and the {\em update gate} $z_t$ be two vectors, which jointly decide which information should be passed from time $t$-$1$ to time $t$.
To realize GRU's goal, the network performs the following calculation, which sets the hidden state at time $t$.
Its final content is based on updating the hidden state in the previous time (the {\em update gate} $z_t$ determines which information should be passed) and on its candidate memory content $\tilde{h}_{t}$:
\begin{equation}
\label{eq:ht}
h_{t}= z_{t} \odot h_{t-1} + \left(1-z_{t}\right) \odot \tilde{h}_{t},
\end{equation}
where $\odot$ is an element-wise multiplication.
Here, $\tilde{h}_{t}$ is defined as:
\begin{equation}
\label{eq:h-tilde}
\tilde{h}_{t}=\tanh \left(W^{(h)} x_{t}+U^{(h)} h_{t-1} \odot r_{t}+b^{(h)}\right).
\end{equation}
That is, when the reset gate is close to $0$, the hidden state ignores the previous hidden state and resets with the current input only.
This effectively allows the hidden state to drop any information that will later be found to be irrelevant.

Finally,  the {\em reset gate} $r_t$ and the {\em update gate} $z_t$ are defined as:
\begin{equation}
\label{eq:z-t}
z_{t}=\sigma\left(W^{(z)} x_{t}+U^{(z)} h_{t-1}+b^{(z)}\right),
\end{equation}
\begin{equation}
\label{eq:rt}
r_{t}=\sigma\left(W^{(r)} x_{t}+U^{(r)} h_{t-1}+b^{(r)}\right),
\end{equation}
where $\sigma$ is a logistic Sigmoid function. 
$W^{(h)}, W^{(z)}, W^{(r)}, U^{(h)}, U^{(z)}$ and $U^{(r)}$ are trainable weight matrices and  $b^{(h)}, b^{(r)}, b^{(r)}$ are trainable bias vectors.
The initial hidden state $h_j$ is set to $0$.

GRU  outperforms a vanilla RNN, due to its ability to both remember the important information along the sequence and to forget unimportant content.
Furthermore, it is capable of processing long sequences, similarly to the {\em Long Short-Term Memory (LSTM)}~\cite{hochreiter1997long}.
Being able to  accumulate information from long sequences is vital for grasping the shape of a 3D model, which usually consists of thousands of vertices. 
We chose GRU over LSTM due to its simplicity and its smaller computational requirements.
For comparison, LSTM would require $16.8M$ trainable parameters in our case, whereas $GRU$ uses $12.7M$.
Furthermore, the inference time is smaller---for instance, a single $100$-steps walk takes $5mSec$ using LSTM and $3mSec$ using GRU.

The third sub-network in Fig.~\ref{fig:NetworkArch} predicts the object class in case of classification,  
or the vertex segment in case of semantic segmentation. 
It consists of a single fully connected (FC) layer on top of the state vector calculated in the previous sub-network.
More details on the architectures \& the implementation are given in Section~\ref{sec:experiments}.

\paragraph{Loss calculation.}
The {\em Softmax} cross entropy loss is used on the output of the third part of the network.
In the case of the classification task, only the last step of the walk is used as input to the loss function, since it accumulates all prior information from the walk. 
In Fig.~\ref{fig:NetworkArch}, this is the bottom-right orange component.

In the case of the segmentation task,
each vertex has its own predicted segment class.
Each of the orange components in Fig.~\ref{fig:NetworkArch} classifies the segment that the respected vertex belongs to. 
Since at the beginning of the walk the results are not trustworthy (as the mesh is not yet well understood), for the loss calculation in the training process we consider the segment class predictions only for the vertices that belong to the second half of the walk.

\section{Applications: Classification \& Segmentation}
\label{sec:applications}
MeshWalker is a general approach, which may be applied to a variety of applications.
We demonstrate its performance for two fundamental tasks in shape analysis: mesh classification and mesh semantic segmentation. 
Our results are compared against the {\em reported} SOTA results for recently-used datasets, hence the methods we compare against vary according to the specific dataset.

\subsection{Mesh classification}
\label{subsec:classification}
Given a mesh, the goal is to classify it into one of pre-defined classes.
For the given mesh we generate multiple random walks.
These walks are run through the trained network. 
For each walk, the network predicts the probability of this mesh to belong to each class.
These prediction vectors are averaged into a single prediction vector.
In practice we use $32$ walks; Section~\ref{sec:experiments} will discuss the robustness of MeshWalker to the number of walks. 

To test our algorithm, we applied our method to three recently-used datasets: SHREC11~\cite{lian2011shape}, engraved cubes ~\cite{hanocka2019meshcnn} and ModelNet40~\cite{wu20153d}, which differ from each other in the number of classes, the number of objects per class, as well as the type of shapes they contain.
As common, the accuracy is defined as the ratio of correctly predicted meshes.

\paragraph{SHREC11.}
This dataset consists of $30$ classes, with 20 examples per class.
Typical classes are camels, cats, glasses, centaurs, hands etc.
Following the setup of~\cite{ezuz2017gwcnn}, we split the objects in each class into $16$ (/$10$)  training examples and $4$  (/$10$) testing examples.

Table~\ref{tbl:shrec11} compares the performance, where each result is the average of the results of $3$ randoms splits (of $16/4$ or of $10/10$). 
When the split is $10$ objects for training and $10$ for testing, the advantage of our method is apparent. 
When $16$ objects are used for training and only $4$ for testing, we get the same accuracy as that of the current state-of-the-art. 
In Section~\ref{subsec:size_training_set} we show that indeed the smaller the training dataset, the more advantageous our approach is.

\begin{table}[htb]
 \caption{{\bf Classification on SHREC11~\cite{lian2011shape}.}
 Split-$16$ and Split-$10$ are the number of training models per class (out of $20$ models in the class).
 In both cases our method achieves state-of-the-art results, yet it is most advantageous for a small training dataset (Split-$10$).
 (We have not found point cloud-based networks that were tested on SHREC11).
 }
\begin{center}
 \begin{tabular}
 {||l l c c||} 
 \hline
 Method & Input & Split-$1$6  & Split-$10$\\ [0.5ex] 
 \hline\hline\hline
 MeshWalker (ours) & Mesh & \textbf{98.6\%} & \textbf{97.1\%} \\ 
 \hline
 MeshCNN~\cite{hanocka2019meshcnn} & Mesh & \textbf{98.6\%} & 91.0\% \\
 \hline
 GWCNN~\cite{ezuz2017gwcnn} & Mesh & 96.6\% & 90.3\% \\
 \hline
 SG~\cite{bronstein2011shape} & Mesh & 70.8\% & 62.6\% \\
 \hline
 \hline
\end{tabular}
\label{tbl:shrec11}
\end{center}
\end{table}

\begin{figure}[htb] 
\centering 
\includegraphics[width=0.48\textwidth]{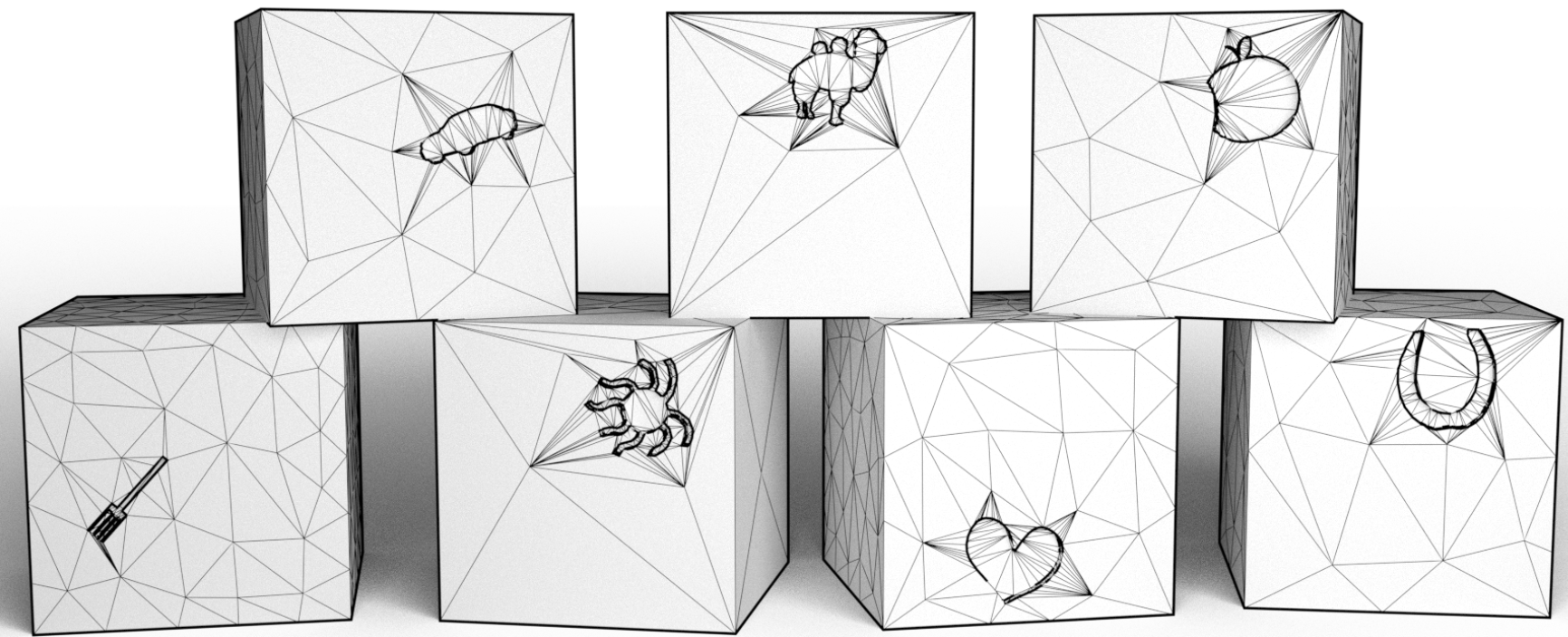}
\caption{{\bf Engraved cubes dataset.} 
This image is courtesy of~\cite{hanocka2019meshcnn}.
}
\label{fig:cube} 
\end{figure}

\paragraph{Cube engraving.}
This dataset contains $4600$ objects, with $3910$/$690$ training/testing split.
Each object is a cube "engraved" with a shape at a random face in a random location, as demonstrated in Fig.~\ref{fig:cube}.
The engraved shape belongs to a dataset of $23$ classes (e.g., car, heart, apple, etc.), each contains roughly $20$ shapes. 
This dataset was created in order to demonstrate that using meshes, rather than point clouds, may be critical for 3D shape analysis.

Table~\ref{tbl:cubes} provides the results.
It demonstrates the benefit of our method over state-of-the-art methods.

\begin{table}[htb]
\caption{{\bf Classification on Cube Engraving~\cite{hanocka2019meshcnn}.} 
Our results outperform those of state-of-the-art algorithms.}
\begin{center}
\begin{tabular}{||l l c||} 
\hline
 Method & Input & accuracy\\ [0.5ex] 
 \hline\hline\hline
 MeshWalker (ours) & Mesh & \textbf{98.6\%} \\ 
 \hline
 MeshCNN~\cite{hanocka2019meshcnn} & Mesh & 92.16\% \\
 \hline\hline
 PointNet++~\cite{qi2017pointnet++} & Point cloud & 64.26\% \\
 \hline
\end{tabular}
\label{tbl:cubes}
\end{center}
\end{table}

\paragraph{ModelNet40.}
This commonly-used dataset contains $12,311$ CAD models from $40$ categories, out of which $9,843$ models are used for training and $2,468$ models are used for testing.
Unlike previous datasets, many of the objects contain multiple components and are not necessarily watertight, making this dataset prohibitive for some mesh-based methods.
However, such models can be handled by MeshWalker since as explained before, if the walk gets into a dead-end during backtracking, it jumps to a new random location.

Table~\ref{tbl:modelnet} shows that our results  outperform those of mesh-based state-of-the-art methods.
We note that without $5$ classes that are cross-labeled (desk/table \& plant/flower-pot/vase) our method's accuracy is $94.4\%$.
The table shows that multi-views approaches are excellent for this dataset. 
This is due to relying on networks that are pre-trained on a large number of images.
However, they might fail for other datasets, such as the engraved cubes, and do not suit other shape analysis tasks, such as semantic segmentation.

\begin{table}[htb]
\caption{{\bf Classification on ModelNet40~\cite{wu20153d}.}
MeshWalker is competitive with other mesh-based methods. 
Multi-view methods  are advantageous for this dataset, possibly due to relying on pre-trained networks for image classification and to naturally handling multiple components and non--watertight models, which characterize many meshes in this dataset.
}
\begin{center}
 \begin{tabular}{||l l c||} 
 \hline
 Method & Input & Accuracy \\ [0.5ex] 
 \hline\hline\hline
 MeshWalker (ours) & mesh & 92.3\% \\ 
 \hline
 MeshNet~\cite{feng2019meshnet} & mesh & 91.9\% \\
 \hline
 SNGC~\cite{haim2019surface} & mesh & 91.6\%  \\
 \hline
 \hline
 KPConv~\cite{thomas2019kpconv} & point cloud & 92.9\% \\
 \hline
 PointNet~\cite{qi2017pointnet} & point cloud & 89.2\%  \\
 \hline
 RS-CNN~\cite{liu2019relation} & point cloud & 93.6\% \\
 \hline
 \hline
 RotationNet~\cite{kanezaki2018rotationnet} & multi-views & \textbf{97.3\%} \\
 \hline
 GVCNN~\cite{feng2018gvcnn} & multi-views & 93.1\% \\
 \hline
 3D2SeqViews~\cite{han20193d2seqviews} & multi-views & 93.4\% \\
 \hline
\end{tabular}
\label{tbl:modelnet}
\end{center}
\end{table}

\begin{figure*}[tb]
\centering
\begin{tabular}{cccc}
\includegraphics[height=0.33\textwidth]{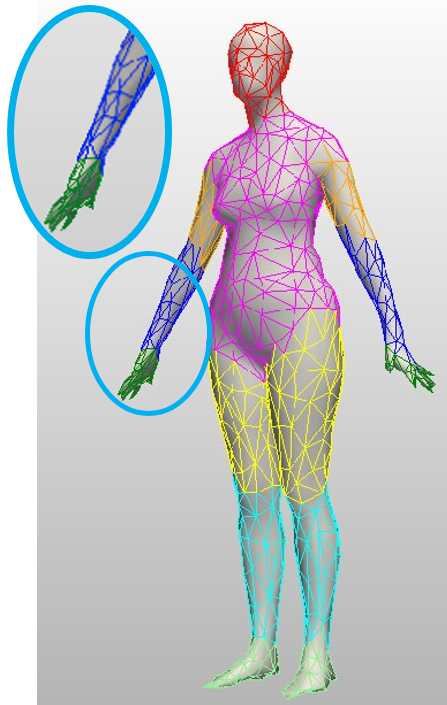}&
\includegraphics[height=0.33\textwidth]{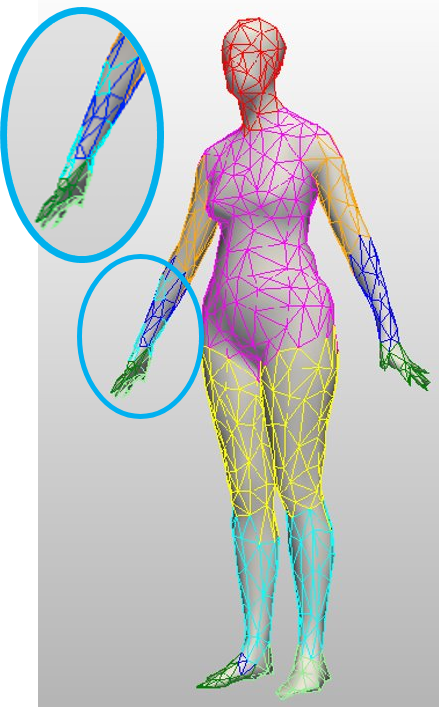}&
\includegraphics[height=0.33\textwidth]{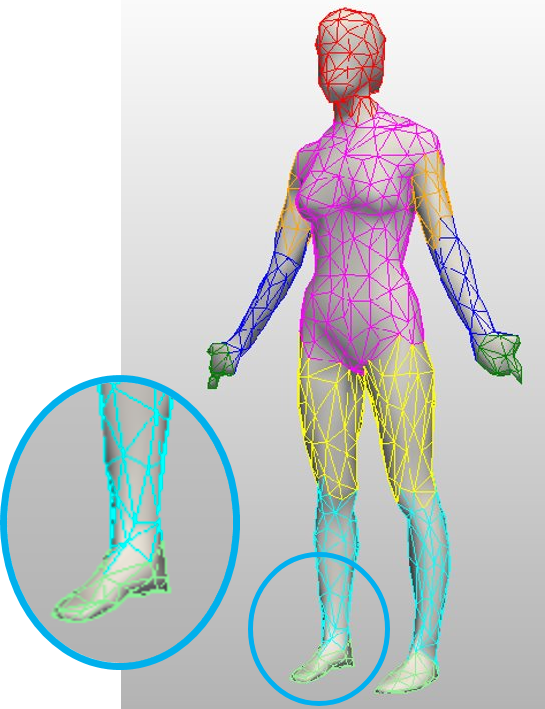}&
\includegraphics[height=0.33\textwidth]{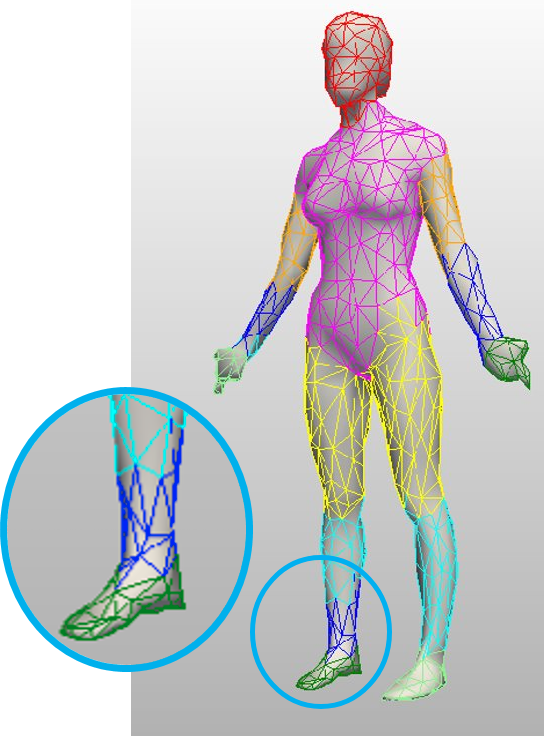}\\
(a) Ours & (b) \cite{hanocka2019meshcnn} & (c) Ours & (d) \cite{hanocka2019meshcnn}
\end{tabular}
\caption{{\bf Qualitative results for human shape segmentation from~\cite{maron2017convolutional}. }
Our system avoids mis-classifications, not mixing lower legs with lower arms or hands with feet.
We note that for most shapes in the dataset, both systems produce equally-good results.
}
\label{fig:human_body_seg_visualization}
\end{figure*}

\subsection{Mesh semantic segmentation}
\label{subsec:segmentation}
Shape segmentation is an important building block for many applications in shape analysis and synthesis.
The goal is to determine, for every vertex, the segment it belongs to.
We tested  MeshWalker on two datasets: COSEG~\cite{wang2012active} and human-body Segmentation~\cite{maron2017convolutional}. 

Given mesh, multiple  random walks are generated (in practice, $32$ $\times$ \# segment classes; see the discussion in Section~\ref{sec:experiments}).
These walks are run through the trained network, which predicts the probabilities of belonging to the segments.
Similarly to the training process, only vertices of the second half of each walk are considered trustworthy.
For each vertex, the predictions of the walks it belongs to are averaged.
Then, as post-processing, we consider the average prediction of the vertex neighbors and add  this average with $0.5$ weight. 
Finally, the prediction for each vertex is the argmax-ed.

Formally, let $\{W\}$ be the set of walks performed on a mesh.
Let $P^i_{v}$ be the vector that is the {\em Softmax} output for vertex $v$ from walk~$i$ (if walk~$i$ does not visit $v$, $P^i_{v}$ is set to a $0$-vector).
Let $v^{ring}$ be the list of the vertices adjacent to $v$ and $N_v$ be the size of this list.
The predicted label, $l_v$  of vertex $v$ is defined as (where $argmax$ finds the maximum vector entry):
\begin{equation}
\label{eqn:lv}
l_v=argmax(\sum_{i \in \{W\}} P^i_{v} + 
 \frac{1}{2N_v} 
\sum_{\tilde{v} \in v^{ring}} 
\sum_{i \in \{W\}} P^i_{{\tilde{v}}}).
\end{equation}

We follow the accuracy measure proposed in~\cite{hanocka2019meshcnn}:
Given the prediction for each edge, the accuracy is defined as the percentage of the correctly-labeled edges, weighted by their length.
Since MeshWalker predicts the segment of the vertices, if the predictions of the endpoints of the edge agree, the edge gets the endpoints' label; otherwise,  the label with the higher prediction is chosen.
The overall accuracy is the average over all meshes.

\paragraph{Human-body segmentation.}
The dataset consists of $370$ training models from SCAPE~\cite{anguelov2005scape}, FAUST~\cite{bogo2014faust}, MIT~\cite{vlasic2008articulated} and Adobe Fuse~\cite{Adobe2016}.
The test set consists of $18$ humans from SHREC'07 \cite{giorgi2007shape} .
The meshes are manually segmented into eight labeled segments according to~\cite{kalogerakis2010learning}.

\begin{table}[t]
\caption{{\bf Human-body segmentation results on~\cite{maron2017convolutional}.}
The accuracy is  calculated on edges of the simplified meshes.
}
\begin{center}
 \begin{tabular}{||l c||} 
 \hline
 Method &  Edge Accuracy\\ [0.5ex] 
 \hline\hline\hline
 MeshWalker  & \textbf{94.8}\% \\ 
 \hline
 MeshCNN & 92.3\%  \\
 \hline
\end{tabular}
\label{tbl:human_body_segmentation_on_edge}
\end{center}
\end{table}

\begin{table}[htb]
\caption{{\bf Human-body segmentation results on~\cite{maron2017convolutional}.} 
The reported results are on the original meshes;
For MeshCNN, the results shown are ours.
Our results outperform those of state-of-the-art algorithms.
}
\begin{center}
 \begin{tabular}{||l l c||} 
 \hline
 Method &  Input & Face\\ [0.5ex] 
   &    & Accuracy\\ [0.5ex] 
 \hline\hline\hline
 MeshWalker (ours)  & Mesh & \textbf{92.7}\% \\ 
 \hline
 MeshCNN~\cite{hanocka2019meshcnn} & Mesh & 89.0\% \\
 \hline
 LRF-Conv~\cite{yang2020continuous} & Mesh & 89.9\% \\
 \hline
SNGC~\cite{haim2019surface} & Mesh & 91.3\%  \\
 \hline
 Toric Cover~\cite{maron2017convolutional} & Mesh & 88.0\%  \\
 \hline
 GCNN~\cite{masci2015geodesic} & Mesh & 86.4\%  \\
 \hline
 MDGCNN & Mesh & 89.5\%  \\
 ~\cite{poulenard2018multi} & &   \\
 \hline
 \hline
 PointNet++~\cite{qi2017pointnet++} & Point cloud & 90.8\% \\
 \hline
 DynGraphCNN~\cite{wang2019dynamic} & Point cloud & 89.7\%  \\
 \hline
\end{tabular}
\label{tbl:human_body_segmentation}
\end{center}
\end{table}

There are two common measures of segmentation results, according to the correct classification of faces~\cite{haim2019surface} or of  edges \cite{hanocka2019meshcnn}.
Tables~\ref{tbl:human_body_segmentation_on_edge} and~\ref{tbl:human_body_segmentation} compare our results to those of previous works, according to the reported measure and the type of objects (simplified or not).
Since our method is trained on simplified meshes, to get results on the original meshes, we apply a simple projection to the original meshes jointly with boundary smoothing, as in~\cite{katz2003hierarchical}.
In both measures, MeshWalker outperforms other methods. 
Fig.~\ref{fig:human_body_seg_visualization} presents qualitative examples where the difference between the resulting segmentations is evident.

\begin{figure*}[htb]
\centering
\begin{tabular}{ccc}
\includegraphics[height=0.17\textwidth]{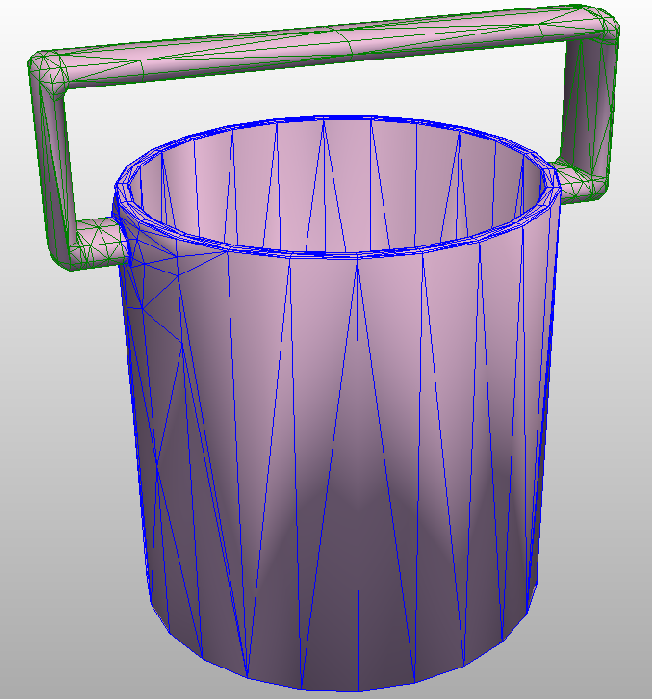}
\includegraphics[height=0.17\textwidth]{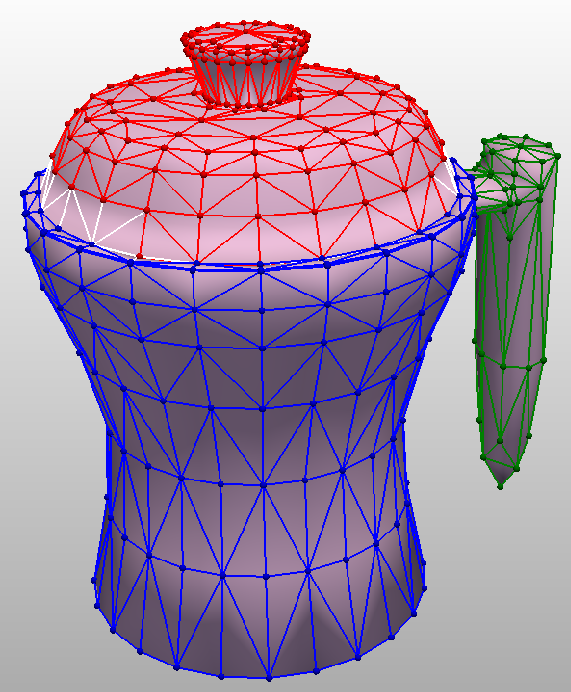}&
\includegraphics[height=0.17\textwidth]{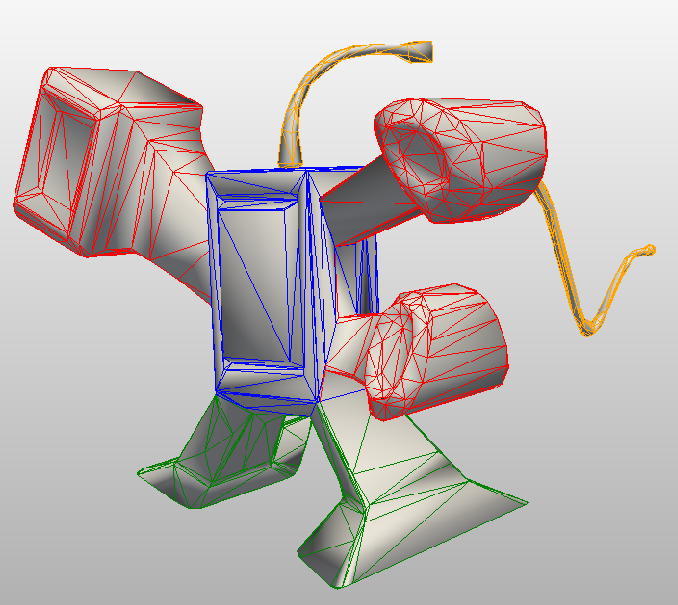}
\includegraphics[height=0.17\textwidth]{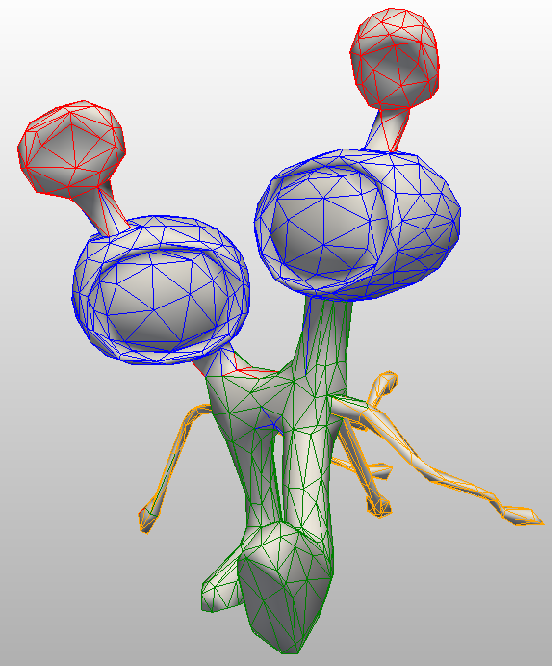}&
\includegraphics[height=0.17\textwidth]{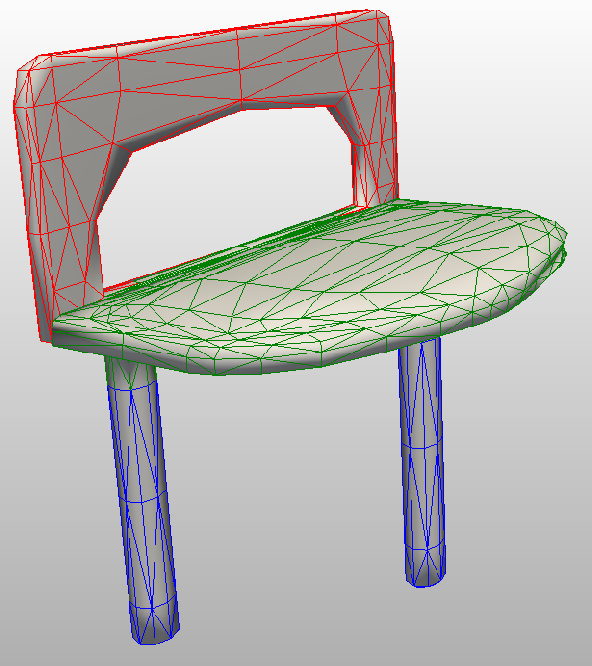}
\includegraphics[height=0.17\textwidth]{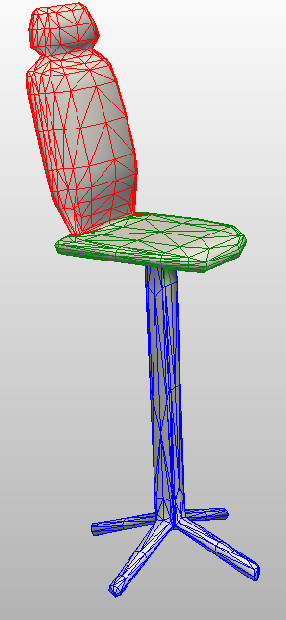}\\
(a) Vases & (b) Aliens & (c) Chairs 
\end{tabular}
\caption{{\bf Qualitative results of segmentation for meshes from COSEG~\cite{wang2012active}.} 
} 
\label{fig:coseg_body_seg_visualization}
\end{figure*}

\paragraph{COSEG segmentation.}
This dataset contains three large classes: aliens, vases and chairs with $200$, $300$ and $400$ shapes,
respectively. 
Each category is split into $85\%$/$15\%$ train/test sets.
Fig.~\ref{fig:coseg_body_seg_visualization} presents some qualitative results, where it can be seen that our method performs very well.
Table~\ref{tbl:coseg} shows the accuracy of our results, where the results of the competitors are reported in~\cite{hanocka2019meshcnn}.
Our method achieves state-of-the-art results for all categories.

\begin{table}[htb]
\caption{{\bf Segmentation results on COSEG~\cite{wang2012active}.}
Our method achieves state-of-the-art results for all categories.
} 
\begin{center}
 \begin{tabular}{||l c c c c||} 
 \hline
 Method & Vases  & Chairs & Telealiens & Mean\\ [0.5ex] 
 \hline\hline\hline
 MeshWalker (ours) & \textbf{98.7}\% & \textbf{99.6}\% & \textbf{99.1}\% & \textbf{99.1}\% \\ 
 \hline
 MeshCNN & 97.3\% & \textbf{99.6}\% & 97.6\% & 98.2\% \\
 \hline\hline
 PointNet++ & 94.7\% & 98.9\% & 79.1\% & 90.9\% \\
 \hline
 PointCNN~\cite{li2018pointcnn} & 96.4\% & 99.3\% & 97.4\% & 97.7\% \\
 \hline
\end{tabular}
\label{tbl:coseg}
\end{center}
\end{table}

\section{Experiments}
\label{sec:experiments}

\subsection{Ablation study}

\paragraph{Size of the training dataset.}
\label{subsec:size_training_set}
How many training models are needed in order to achieve good performance?
In the 3D case this question is especially important, since creating a dataset is costly.
Table~\ref{tbl:train_size_alien_coseg} shows the accuracy of our model for the {\em COSEG} dataset, when trained on different dataset sizes.
As expected, the larger the dataset, the better the results.
However, even when using only $4$ shapes for training, the results are pretty good ($80.5\%$). 
This outstanding result can be explained by the fact that 
we can produce many random walks for each mesh, hence the actual number of training examples is large. 
This result is consistent across all categories and datasets.
Table~\ref{tbl:train_size_human_body} shows a similar result for the  {\em human-body segmentation} dataset.

\begin{table}[htb]
\caption{{\bf Analysis of the training dataset size (COSEG segmentation).} 
"Full" training is $170$, $255$ and $240$ shapes for tele-aliens, vases and chairs, respectively.
As expected, the larger the dataset, the better the results.
However, even if the training dataset is very small, our results are good.
}
\begin{center}
 \begin{tabular}{||c c c c c||} 
 \hline
 \# training shapes & Vases & Chairs & Tele-aliens & Mean \\ [0.5ex] 
 \hline\hline\hline
 Full & $98.7\%$ & $99.6\%$ & $99.1\%$ & $99.1\%$ \\ 
 \hline\hline
 32         & $95.3\%$ & $98.5\%$ & $94.2\%$ & $96.0\%$ \\
 \hline
 16         & $93.6\%$ & $93.4\%$ & $92.4\%$ & $93.1\%$ \\
 \hline
 8          & $83.7\%$ & $87.7\%$ & $86.7\%$ & $86.0\%$ \\
 \hline
 4          & $77.5\%$ & $83.7\%$ & $80.4\%$ & $80.5\%$ \\
 \hline
 2          & $67.3\%$ & $78.4\%$ & $69.7\%$ & $71.8\%$ \\
 \hline
 1          & $60.9\%$ & $59.9\%$ & $40.6\%$ & $53.8\%$ \\
 \hline
\end{tabular}
\label{tbl:train_size_alien_coseg}
\end{center}
\end{table}

\begin{table}[htb]
\caption{{\bf Analysis of the training dataset size (human-body segmentation).}
As before, the performance of our method degrades gracefully with the size of the training set.
We note that the results of MeshCNN are not reported in their paper, but rather the results of  new runs of their system.
}
\begin{center}
 \begin{tabular}{||c c c||} 
 \hline
 \# training shapes & MeshWalker & MeshCNN \\ [0.5ex] 
   & (ours) & \cite{hanocka2019meshcnn} \\
\hline\hline\hline
 381 (full) & $\textbf{94.8\%}$ & $92.3\%$ \\ 
 \hline\hline
 16 & $\textbf{92.0\%}$ & $55.7\%$ \\
 \hline
 4 & $\textbf{84.3\%}$ & $48.3\%$ \\
 \hline
 2 & $\textbf{80.8\%}$ & $42.4\%$ \\
 \hline
\end{tabular}
\label{tbl:train_size_human_body}
\end{center}
\end{table}

\begin{figure}[tb] 
\centering 
\includegraphics[width=0.45\textwidth]
{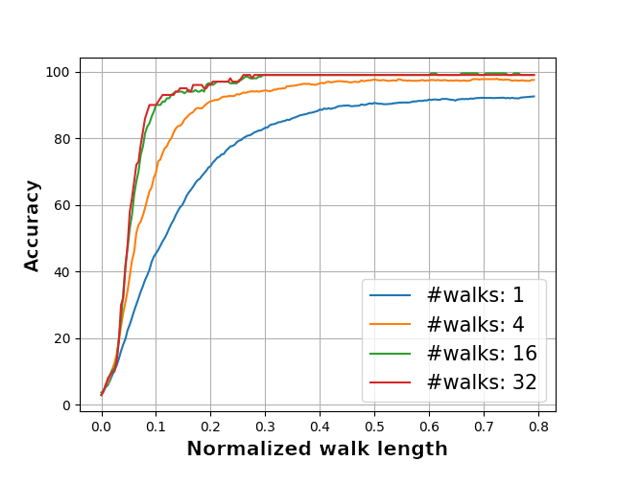} 
\caption{{\bf Walk length analysis.} 
The accuracy increases with walk length, for classification on {\em SHREC11}.
Here, the $X$ axis is number of vertices along the walk, normalized by number of mesh vertices.
This figure illustrates trade-off between the number of walks we use and the length of these walks.
As the walk begins, using many walks is not beneficial since the RNN has not accumulated enough information yet. However, after e.g.~0.3V, two walks are better than a single 0.6V-length walk. This is because they explore different mesh regions. 
}
\label{fig:walk_length_analysis} 
\end{figure}

\paragraph{Walk length.}
Fig.~\ref{fig:teaser} has shown that the accuracy of our method depends on the walk length.
What would be an ideal length for our system to "understand" a shape?
Fig.~\ref{fig:walk_length_analysis} analyzes the influence of the length on the task of classification for {\em SHREC11}.
As expected, the accuracy increases with length.
However, it can be seen that when we use at least $16$ walks per mesh, a walk whose length is $0.15V$ suffices to get excellent results.
Furthermore, there is a trade-off between the number of walks we use and the length of these walks.
Though the exact length depends both on the task in hand and on the dataset, this correlation is consistent across datasets and tasks.

\paragraph{Number of walks.}
How many walks are needed at inference time?
Table~\ref{tbl:accuracy_per_n_walks} shows that the more walks, the better the accuracy. However, even very few walks result in very good accuracy.
In particular, on SHREC11, even with a single walk the accuracy is $90.8\%$.
For the Engraved-Cubes dataset, more walks are needed, since the model is engraved on a single cube facet, which certain walks might not get to.
Even in this difficult case, $4$ walks already achieve $92.1\%$ accuracy.
We note that the STD is between $2.5\%$ for a single walk to $0.4\%$ for $32$ walks.
As expected, the more walks used, the more stable the results are and the smaller the STD is.

\begin{table}[tb]
\caption{{\bf Number of walks analysis.}
The accuracy improves with the number of walks per shape (demonstrated on $2$ datasets). 
}
\begin{center}
 \begin{tabular}{||c c c||} 
 \hline
 \# Walks & SHREC11 Acc & Eng.Cubes Acc\\ [0.5ex] 
 \hline\hline\hline
 $32$ & $98.3\%$ & $97.6$\%  \\ 
 \hline
 $16$ & $97.8\%$ & $97.4$\%   \\ 
 \hline
 $8$ & $97.8\%$ & $95.3$\%   \\ 
 \hline
 $4$ & $95.5\%$ & $92.1$\%   \\ 
 \hline
 $2$ & $95.0\%$ & $84.8$\%   \\ 
 \hline
 $1$ & $90.8\%$ & $77.1$\%   \\ 
 \hline
  \end{tabular}
\label{tbl:accuracy_per_n_walks}
\end{center}
\end{table}

\paragraph{Robustness.}
We use various rotations within data augmentation, hence robustness to orientations.
In particular, to test the robustness to rotation, we rotated the models in the Human-body segmentation dataset and in SHREC11 classification dataset $36$ times for each axis, by increments of $10^\circ$.
For each of these rotated versions of the datasets we applied the same testing as before.
For both datasets, there was no difference in the results.
Furthermore, the meshes are normalized, hence robustness to scaling. 

Our approach is inherently robust to different triangulations, as random walks (representing the same mesh) may vary greatly anyhow. 
Specifically, we generated a modified version of the COSEG segmentation dataset by randomly perturbing $30\%$ of the vertex positions, realized as a shift towards a random vertex in its $1$-ring. 
The performance degradation is less than $0.1\%$.

\begin{figure*}
\centering
\begin{tabular}{cccccc}
\includegraphics[width=0.15\textwidth]{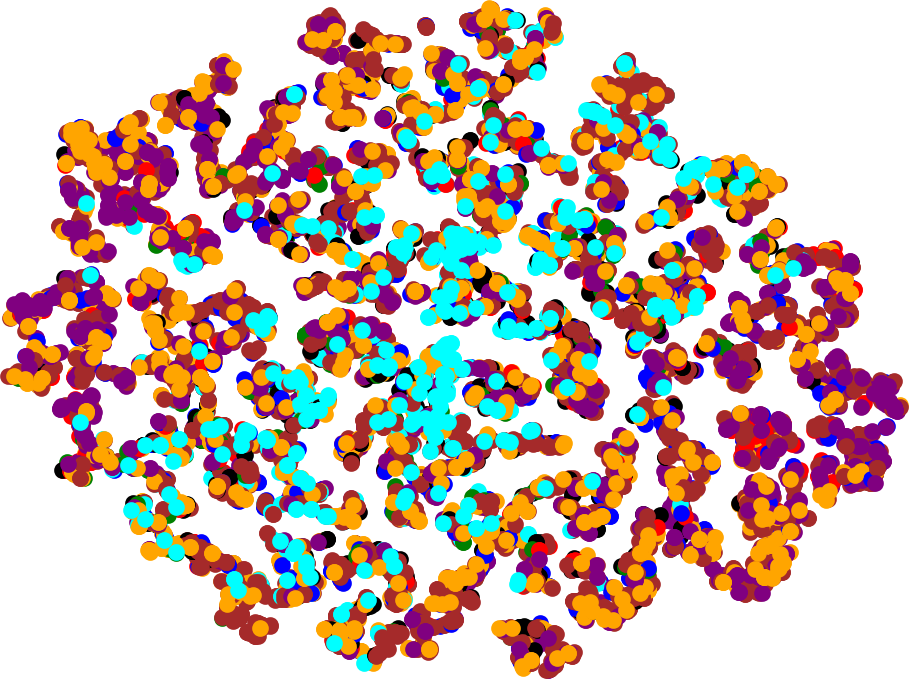}&
\includegraphics[width=0.15\textwidth]{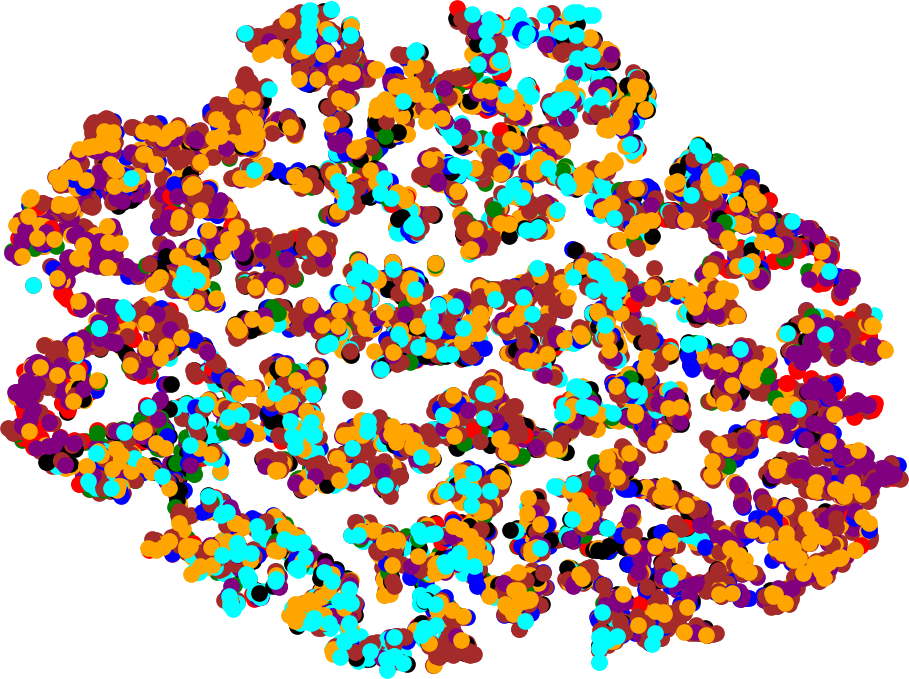}&
\includegraphics[width=0.15\textwidth]{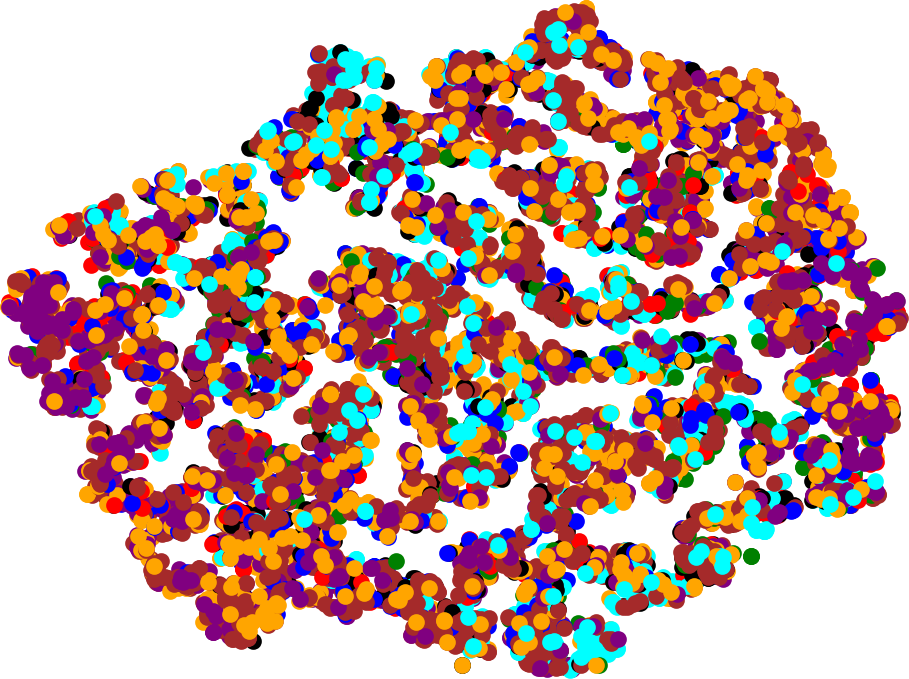}&
\includegraphics[width=0.15\textwidth]{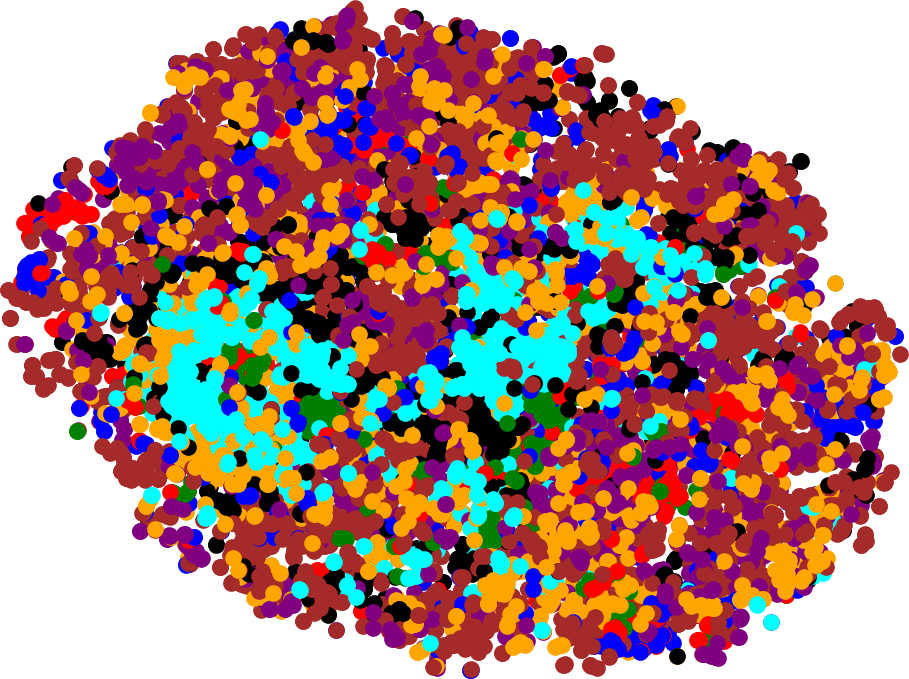}&
\includegraphics[width=0.15\textwidth]{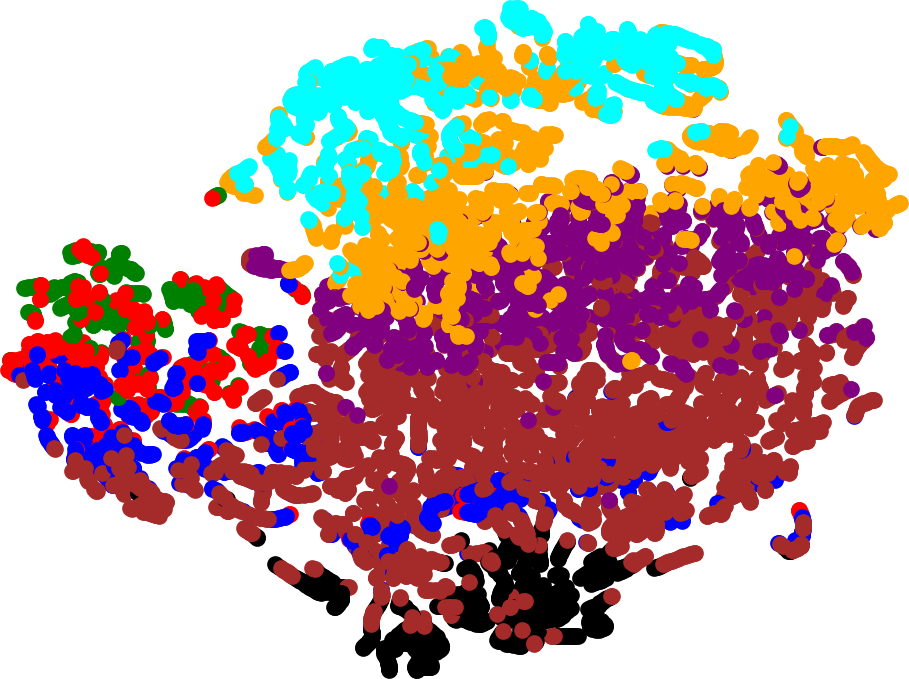}&
\includegraphics[width=0.15\textwidth]{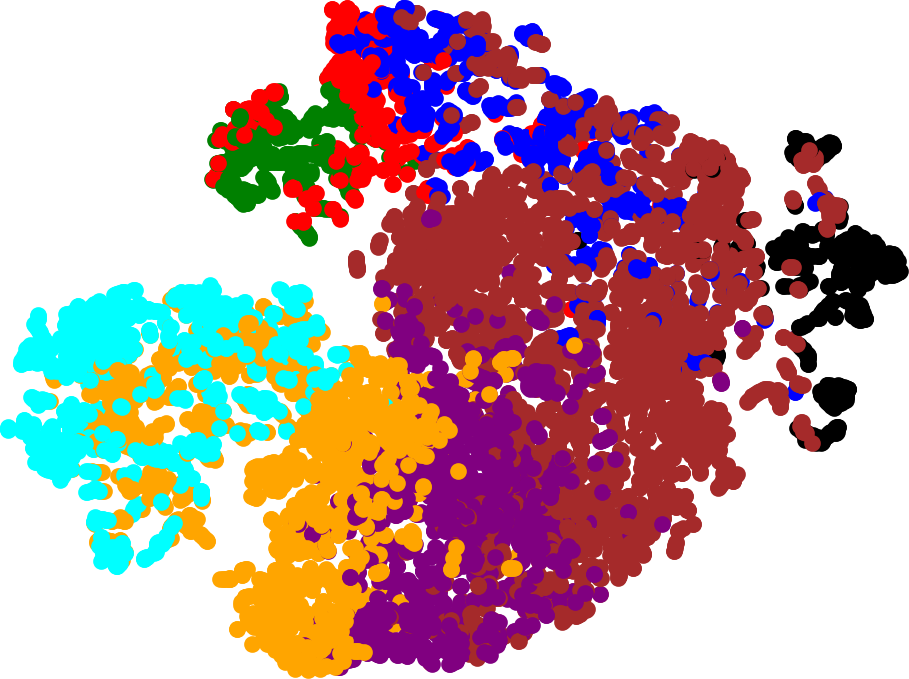}\\
(a) input &
(b) FC1 &
(c) FC2 &
(d) GRU1&
(e) GRU2 &
(f) GRU3
\end{tabular}
\caption{{\bf t-SNE of the internal layers.}
This is a visualization of the output of the different layers for the human-body segmentation task.
It can be seen how the semantic meaning of the layers' output starts to evolve after the first GRU layer and gets better in the next two layers.
} 
\label{fig:t_sne_all_layers}
\end{figure*}

\subsection{Implementation}
\label{subsec:implementation}

\paragraph{Mesh pre-processing: simplification \& data augmentation.}
\label{subsec:preprocessing}
All the meshes used for training are first simplified into roughly the same number of faces~\cite{garland1997surface, hoppe1997view} ({\em MeshProcessing} procedure in Algorithm~\ref{alg:MeshWalkerTraining}).
Simplification is analogous to the initial resizing of images.
It reduces the network capacity required for training. 
Moreover, we could use several simplifications for each mesh as a form of data augmentation for training and for testing. 
For instance, for ModelNet40 we use $1K$, $2K$ and $4K$ faces.
The meshes are normalized into a unit sphere, if necessary.

In addition, we augment the training data and add diversity by rotating the models. 
As part of batch preparation, each model is randomly rotated in each  axis prior to each training iteration.

\paragraph{t-SNE analysis.}
Does the network produce meaningful features?  
Fig.~\ref{fig:t_sne_all_layers} opens the network's "black box" and shows the t-SNE  projection to 2D of the multi-dimensional features after each stage of our learning framework, applied to the human-body segmentation task.
Each feature vector is colored by its correct label. 

In the input layer all the classes are mixed together. 
The same behavior is noticed after the first two fully-connected layers, since no information is shared between the vertices up to this stage. 
In the next three GRU layers, semantic meaning evolves:
The features are structured as we get deeper in the network.
In the last RNN layer the features are meaningful, as the clusters are evident. 
This visualization demonstrates the importance of the RNN hierarchy.

\begin{figure}[tb] 
\centering 
\includegraphics[width=0.35\textwidth]{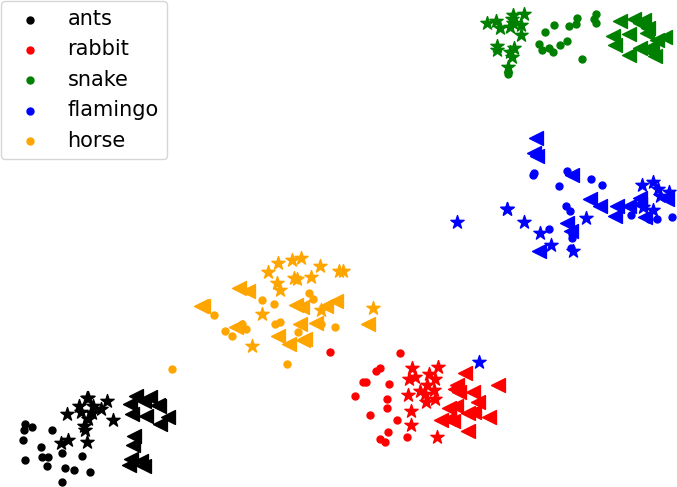} 
\caption{
{\bf t-SNE analysis for classification.}
This figure shows feature hierarchy:
Meshes that belong to the same category (indicated by the color) are clustered together.
Furthermore, walks that belong to the same mesh  (indicated by the shape of the 2D point) are also clustered.
}
\label{fig:t-sne_shrec} 
\end{figure}

Fig.~\ref{fig:t-sne_shrec} reveals another invaluable property of our walks.
It shows the t-SNE visualization of walks for classification of objects from $5$ categories of  SHREC11.
Each feature vector is colored by its correct label; its shape (rectangle, triangle etc) represents the object the walk belongs to.
Not only clusters of shapes from the same category clearly emerge, but also walks that belong to the same object are grouped together!
This is another indication to the quality of our proposed features.

\paragraph{Computation time.}
Training takes between $5$ hours (for classification on SHREC11) to $12$ hours (for segmentation on human-body), using GTX 1080 TI graphics card.
At inference, a $100$-step walk, which is typical for  SHREC11, takes about $4$ milliseconds.
When we use $32$ walks per shape, the running time would be $128$ milliseconds. 
Remeshing takes e.g. $4.6$ seconds from $400K$ faces to $1.5K$ or $0.85$ from $100K$ face  to $1.5K$ faces.
We note that our method is easy to parallelize, as every walk could be processed on a different processor, which is yet another benefit of our approach.

\paragraph{Training configurations.}
We implemented our network using {\em TensorFlow V2}. 
The network architecture is given in Table~\ref{tbl:configuration}.
The source code is available on
\url{"https://github.com/AlonLahav/MeshWalker"}.

\begin{table}[htb]
\caption{{\bf Training configuration}}
\begin{center}
 \begin{tabular}{||l c||} 
 \hline
 Layer & Output Dimension \\ [0.5ex] 
 \hline\hline\hline
 Vertex description & $3$  \\ 
 Fully Connected & $128$  \\ 
 Instance Normalization & $128$ \\ 
 ReLU & $128$ \\
 Fully Connected & $256$  \\ 
 Instance Normalization & $256$ \\ 
 ReLU & $256$ \\
 GRU & $1024$ \\
 GRU & $1024$ \\
 GRU & $512$ \\
 Fully Connected & \# of classes  \\ 
 \hline
\end{tabular}
\label{tbl:configuration}
\end{center}
\end{table}

Optimization: 
To update the network weights, we use Adam optimizer~\cite{kingma2014adam}. 
The learning rate is set in a cyclic way, as suggested  by~\cite{smith2017cyclical}.
The initial and the maximum learning rates are set to $10^{-6}$ and $5 \cdot 10^{-4}$ respectively.
The cycle size is $20k$ iterations.

Batch strategy:
Walks are grouped into batches of $32$ walks each.
For mesh classification, the walks are generated from different meshes, whereas for semantic segmentation each batch is composed of $4$ walks on $8$ meshes.

Training iterations:
We train for $60$k, $60$k, $460$k, $200$k, $200$k iterations for SHREC11, COSEG, human-body segmentation, engraved-cubes and ModelNet40 datasets, respectively.
This is so since for the loss to converge fast, many of the walks should cover the salient parts of the shape, which distinguish it from other classes/segments.
When this is not the case, more iterations are needed in order for the few meaningful walks to influence the loss.
This is the case for instance in the engraved cubes dataset, where the salient information lies on a single facet.

\subsection{Limitations}
\label{subsec:limitations}

\begin{figure}[htb]
\centering
\begin{tabular}{ccc}
\includegraphics[height=0.22\textwidth]{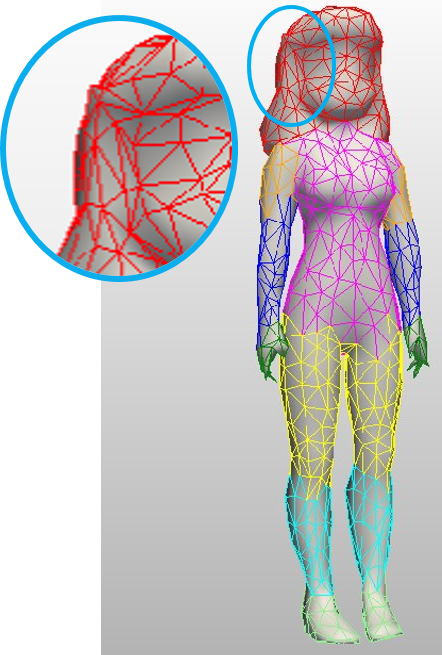}&
\includegraphics[height=0.22\textwidth]{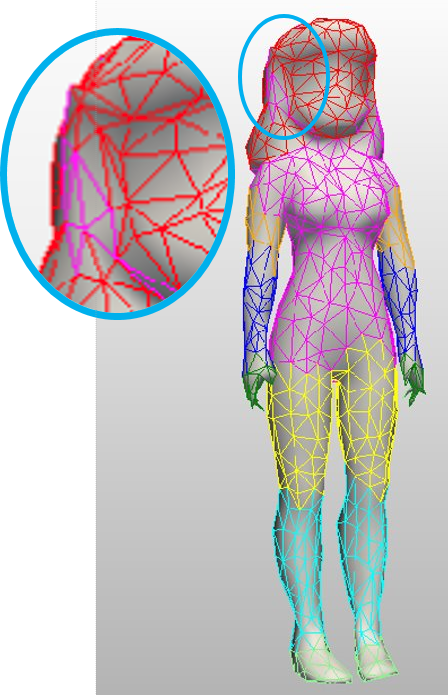}&
\hspace{-0.1in}\includegraphics[height=0.22\textwidth]{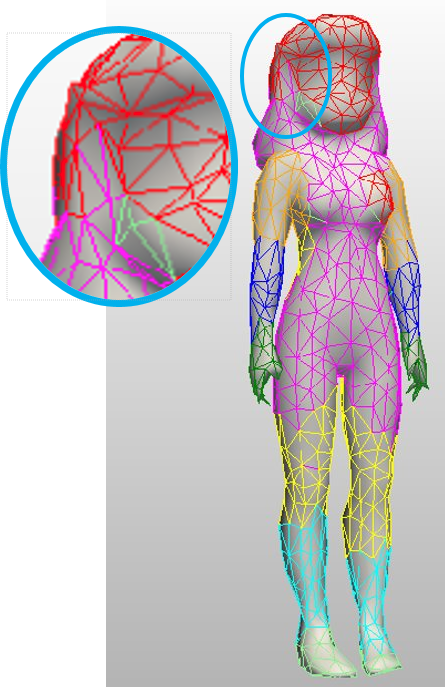}\\
(a) Ground truth & (b) Ours & \hspace{-0.1in}(c) \cite{hanocka2019meshcnn}
\end{tabular}
\caption{{\bf Limitation. }
Our algorithm fails to classify the hair due to not having sufficient similar shapes in the dataset.
}
\label{fig:limitation_segmentation}
\end{figure}

Fig.~\ref{fig:limitation_segmentation} shows a failure of our algorithm, where parts of the hair were wrongly classified as  a torso.
 This is the case since the training data does not contain enough models with hair to learn from.
In general, learning-based algorithms rely on good training data, which is not always available.

Another limitation is handling large meshes.
The latter require long walks, which in turn might lead to run-time and memory issues.
In this paper, this is solved by simplifying the meshes and then projecting the segmentation results onto the original meshes.
(For classification, this is not a concern, as simplified meshes may be used).

\section{Conclusion}
\label{sec:conclusion}
This paper has introduced a novel approach for representing meshes within deep learning schemes. 
The key idea is to represent the mesh by random walks on its surface, which intuitively explore the shape of the mesh.
Since walks are described by the order of visiting mesh vertices, they suit deep learning.

Utilizing this representation, the paper has proposed an end-to-end learning framework, termed {\em MeshWalker}.
The random walks are fed into a Recurrent Neural Network (RNN), that "remembers" the walk’s history (i.e. the geometry of the mesh).
Prior works indicated that RNNs are unsuitable for point clouds due to both the unordered nature of the data and the number of vertices used to represent a shape. Surprisingly, we have shown that RNNs work extremely well for meshes, through the concept of random walks. 

Our approach is general, yet simple.
It has several additional benefits.
Most notably, it works well even for extremely small datasets. e.g.  even $4$ meshes per class suffice  to get good results. 
In addition, the meshes are not require to be watertight or to consist of a single component (as demonstrated by ModelNet40~\cite{wu20153d}); some other mesh-based approaches impose these conditions and require the meshes to be manifolds.

Last but not least, the power of this approach has been demonstrated for two key tasks in shape analysis: mesh classification and mesh semantic segmentation. 
In both cases, we present  state-of-the-art  results.

An interesting question for future work is whether there are optimal walks for meshes, rather than random walks.
For instance, are there good starting points of walks?
Additionally, reinforcement learning could be utilized to learn good walks.
Exploring other applications, such as shape correspondence,
is another intriguing future direction.
Another interesting practical future work would be to work on the mesh as is, without simplification as pre-processing.

\begin{acks}
  We gratefully acknowledge the support of the Israel Science Foundation (ISF) 1083/18 amd PMRI -- Peter Munk Research Institute -- Technion.
\end{acks}

\bibliographystyle{ACM-Reference-Format}
\bibliography{bib}


\begin{thebibliography}{115}


\ifx \showCODEN    \undefined \def \showCODEN     #1{\unskip}     \fi
\ifx \showDOI      \undefined \def \showDOI       #1{#1}\fi
\ifx \showISBNx    \undefined \def \showISBNx     #1{\unskip}     \fi
\ifx \showISBNxiii \undefined \def \showISBNxiii  #1{\unskip}     \fi
\ifx \showISSN     \undefined \def \showISSN      #1{\unskip}     \fi
\ifx \showLCCN     \undefined \def \showLCCN      #1{\unskip}     \fi
\ifx \shownote     \undefined \def \shownote      #1{#1}          \fi
\ifx \showarticletitle \undefined \def \showarticletitle #1{#1}   \fi
\ifx \showURL      \undefined \def \showURL       {\relax}        \fi
\providecommand\bibfield[2]{#2}
\providecommand\bibinfo[2]{#2}
\providecommand\natexlab[1]{#1}
\providecommand\showeprint[2][]{arXiv:#2}

\bibitem[\protect\citeauthoryear{Adobe}{Adobe}{2016}]%
        {Adobe2016}
\bibfield{author}{\bibinfo{person}{Adobe}.} \bibinfo{year}{2016}\natexlab{}.
\newblock \bibinfo{title}{Adobe Fuse 3D Characters}.
\newblock \bibinfo{howpublished}{\url{https://www.mixamo.com}}.
\newblock


\bibitem[\protect\citeauthoryear{Alpert and Yao}{Alpert and Yao}{1995}]%
        {alpert1995spectral}
\bibfield{author}{\bibinfo{person}{Charles~J Alpert} {and}
  \bibinfo{person}{So-Zen Yao}.} \bibinfo{year}{1995}\natexlab{}.
\newblock \showarticletitle{Spectral partitioning: the more eigenvectors, the
  better}. In \bibinfo{booktitle}{\emph{Proceedings of the 32nd annual ACM/IEEE
  Design Automation Conference}}. \bibinfo{pages}{195--200}.
\newblock


\bibitem[\protect\citeauthoryear{Anguelov, Srinivasan, Koller, Thrun, Rodgers,
  and Davis}{Anguelov et~al\mbox{.}}{2005}]%
        {anguelov2005scape}
\bibfield{author}{\bibinfo{person}{Dragomir Anguelov}, \bibinfo{person}{Praveen
  Srinivasan}, \bibinfo{person}{Daphne Koller}, \bibinfo{person}{Sebastian
  Thrun}, \bibinfo{person}{Jim Rodgers}, {and} \bibinfo{person}{James Davis}.}
  \bibinfo{year}{2005}\natexlab{}.
\newblock \showarticletitle{SCAPE: shape completion and animation of people}.
\newblock In \bibinfo{booktitle}{\emph{ACM SIGGRAPH 2005 Papers}}.
  \bibinfo{pages}{408--416}.
\newblock


\bibitem[\protect\citeauthoryear{Attene, Falcidieno, and Spagnuolo}{Attene
  et~al\mbox{.}}{2006}]%
        {attene2006hierarchical}
\bibfield{author}{\bibinfo{person}{Marco Attene}, \bibinfo{person}{Bianca
  Falcidieno}, {and} \bibinfo{person}{Michela Spagnuolo}.}
  \bibinfo{year}{2006}\natexlab{}.
\newblock \showarticletitle{Hierarchical mesh segmentation based on fitting
  primitives}.
\newblock \bibinfo{journal}{\emph{The Visual Computer}} \bibinfo{volume}{22},
  \bibinfo{number}{3} (\bibinfo{year}{2006}), \bibinfo{pages}{181--193}.
\newblock


\bibitem[\protect\citeauthoryear{{Attene}, {Katz}, {Mortara}, {Patane},
  {Spagnuolo}, and {Tal}}{{Attene} et~al\mbox{.}}{2006}]%
        {Attene_06}
\bibfield{author}{\bibinfo{person}{M. {Attene}}, \bibinfo{person}{S. {Katz}},
  \bibinfo{person}{M. {Mortara}}, \bibinfo{person}{G. {Patane}},
  \bibinfo{person}{M. {Spagnuolo}}, {and} \bibinfo{person}{A. {Tal}}.}
  \bibinfo{year}{2006}\natexlab{}.
\newblock \showarticletitle{Mesh Segmentation - A Comparative Study}. In
  \bibinfo{booktitle}{\emph{IEEE International Conference on Shape Modeling and
  Applications 2006 (SMI'06)}}. \bibinfo{pages}{7--7}.
\newblock


\bibitem[\protect\citeauthoryear{Atzmon, Maron, and Lipman}{Atzmon
  et~al\mbox{.}}{2018}]%
        {atzmon2018point}
\bibfield{author}{\bibinfo{person}{Matan Atzmon}, \bibinfo{person}{Haggai
  Maron}, {and} \bibinfo{person}{Yaron Lipman}.}
  \bibinfo{year}{2018}\natexlab{}.
\newblock \showarticletitle{Point convolutional neural networks by extension
  operators}.
\newblock \bibinfo{journal}{\emph{arXiv preprint arXiv:1803.10091}}
  (\bibinfo{year}{2018}).
\newblock


\bibitem[\protect\citeauthoryear{Bai, Bai, Zhou, Zhang, and Jan~Latecki}{Bai
  et~al\mbox{.}}{2016}]%
        {bai2016gift}
\bibfield{author}{\bibinfo{person}{Song Bai}, \bibinfo{person}{Xiang Bai},
  \bibinfo{person}{Zhichao Zhou}, \bibinfo{person}{Zhaoxiang Zhang}, {and}
  \bibinfo{person}{Longin Jan~Latecki}.} \bibinfo{year}{2016}\natexlab{}.
\newblock \showarticletitle{Gift: A real-time and scalable 3d shape search
  engine}. In \bibinfo{booktitle}{\emph{Proceedings of the IEEE conference on
  computer vision and pattern recognition}}. \bibinfo{pages}{5023--5032}.
\newblock


\bibitem[\protect\citeauthoryear{Ben-Shabat, Lindenbaum, and
  Fischer}{Ben-Shabat et~al\mbox{.}}{2018}]%
        {ben20183dmfv}
\bibfield{author}{\bibinfo{person}{Yizhak Ben-Shabat}, \bibinfo{person}{Michael
  Lindenbaum}, {and} \bibinfo{person}{Anath Fischer}.}
  \bibinfo{year}{2018}\natexlab{}.
\newblock \showarticletitle{3dmfv: Three-dimensional point cloud classification
  in real-time using convolutional neural networks}.
\newblock \bibinfo{journal}{\emph{IEEE Robotics and Automation Letters}}
  \bibinfo{volume}{3}, \bibinfo{number}{4} (\bibinfo{year}{2018}),
  \bibinfo{pages}{3145--3152}.
\newblock


\bibitem[\protect\citeauthoryear{Bogo, Romero, Loper, and Black}{Bogo
  et~al\mbox{.}}{2014}]%
        {bogo2014faust}
\bibfield{author}{\bibinfo{person}{Federica Bogo}, \bibinfo{person}{Javier
  Romero}, \bibinfo{person}{Matthew Loper}, {and} \bibinfo{person}{Michael~J
  Black}.} \bibinfo{year}{2014}\natexlab{}.
\newblock \showarticletitle{FAUST: Dataset and evaluation for 3D mesh
  registration}. In \bibinfo{booktitle}{\emph{Proceedings of the IEEE
  Conference on Computer Vision and Pattern Recognition}}.
  \bibinfo{pages}{3794--3801}.
\newblock


\bibitem[\protect\citeauthoryear{Boscaini, Masci, Rodol{\`a}, and
  Bronstein}{Boscaini et~al\mbox{.}}{2016}]%
        {boscaini2016learning}
\bibfield{author}{\bibinfo{person}{Davide Boscaini}, \bibinfo{person}{Jonathan
  Masci}, \bibinfo{person}{Emanuele Rodol{\`a}}, {and} \bibinfo{person}{Michael
  Bronstein}.} \bibinfo{year}{2016}\natexlab{}.
\newblock \showarticletitle{Learning shape correspondence with anisotropic
  convolutional neural networks}. In \bibinfo{booktitle}{\emph{Advances in
  neural information processing systems}}. \bibinfo{pages}{3189--3197}.
\newblock


\bibitem[\protect\citeauthoryear{Boulch, Le~Saux, and Audebert}{Boulch
  et~al\mbox{.}}{2017}]%
        {boulch2017unstructured}
\bibfield{author}{\bibinfo{person}{Alexandre Boulch}, \bibinfo{person}{Bertrand
  Le~Saux}, {and} \bibinfo{person}{Nicolas Audebert}.}
  \bibinfo{year}{2017}\natexlab{}.
\newblock \showarticletitle{Unstructured Point Cloud Semantic Labeling Using
  Deep Segmentation Networks.}
\newblock \bibinfo{journal}{\emph{3DOR}}  \bibinfo{volume}{2}
  (\bibinfo{year}{2017}), \bibinfo{pages}{7}.
\newblock


\bibitem[\protect\citeauthoryear{Brock, Lim, Ritchie, and Weston}{Brock
  et~al\mbox{.}}{2016}]%
        {brock2016generative}
\bibfield{author}{\bibinfo{person}{Andrew Brock}, \bibinfo{person}{Theodore
  Lim}, \bibinfo{person}{James~M Ritchie}, {and} \bibinfo{person}{Nick
  Weston}.} \bibinfo{year}{2016}\natexlab{}.
\newblock \showarticletitle{Generative and discriminative voxel modeling with
  convolutional neural networks}.
\newblock \bibinfo{journal}{\emph{arXiv preprint arXiv:1608.04236}}
  (\bibinfo{year}{2016}).
\newblock


\bibitem[\protect\citeauthoryear{Bronstein, Bronstein, Guibas, and
  Ovsjanikov}{Bronstein et~al\mbox{.}}{2011}]%
        {bronstein2011shape}
\bibfield{author}{\bibinfo{person}{Alexander~M Bronstein},
  \bibinfo{person}{Michael~M Bronstein}, \bibinfo{person}{Leonidas~J Guibas},
  {and} \bibinfo{person}{Maks Ovsjanikov}.} \bibinfo{year}{2011}\natexlab{}.
\newblock \showarticletitle{Shape google: Geometric words and expressions for
  invariant shape retrieval}.
\newblock \bibinfo{journal}{\emph{ACM Transactions on Graphics (TOG)}}
  \bibinfo{volume}{30}, \bibinfo{number}{1} (\bibinfo{year}{2011}),
  \bibinfo{pages}{1--20}.
\newblock


\bibitem[\protect\citeauthoryear{Bronstein, Bronstein, and Kimmel}{Bronstein
  et~al\mbox{.}}{2006}]%
        {bronstein2006efficient}
\bibfield{author}{\bibinfo{person}{Alexander~M Bronstein},
  \bibinfo{person}{Michael~M Bronstein}, {and} \bibinfo{person}{Ron Kimmel}.}
  \bibinfo{year}{2006}\natexlab{}.
\newblock \showarticletitle{Efficient computation of isometry-invariant
  distances between surfaces}.
\newblock \bibinfo{journal}{\emph{SIAM Journal on Scientific Computing}}
  \bibinfo{volume}{28}, \bibinfo{number}{5} (\bibinfo{year}{2006}),
  \bibinfo{pages}{1812--1836}.
\newblock


\bibitem[\protect\citeauthoryear{Chazelle, Dobkin, Shouraboura, and
  Tal}{Chazelle et~al\mbox{.}}{1997}]%
        {chazelle1997strategies}
\bibfield{author}{\bibinfo{person}{Bernard Chazelle}, \bibinfo{person}{David~P
  Dobkin}, \bibinfo{person}{Nadia Shouraboura}, {and} \bibinfo{person}{Ayellet
  Tal}.} \bibinfo{year}{1997}\natexlab{}.
\newblock \showarticletitle{Strategies for polyhedral surface decomposition: an
  experimental study}.
\newblock \bibinfo{journal}{\emph{Computational Geometry}} \bibinfo{volume}{7},
  \bibinfo{number}{5-6} (\bibinfo{year}{1997}), \bibinfo{pages}{327--342}.
\newblock


\bibitem[\protect\citeauthoryear{Cho, Van~Merri{\"e}nboer, Gulcehre, Bahdanau,
  Bougares, Schwenk, and Bengio}{Cho et~al\mbox{.}}{2014}]%
        {cho2014learning}
\bibfield{author}{\bibinfo{person}{Kyunghyun Cho}, \bibinfo{person}{Bart
  Van~Merri{\"e}nboer}, \bibinfo{person}{Caglar Gulcehre},
  \bibinfo{person}{Dzmitry Bahdanau}, \bibinfo{person}{Fethi Bougares},
  \bibinfo{person}{Holger Schwenk}, {and} \bibinfo{person}{Yoshua Bengio}.}
  \bibinfo{year}{2014}\natexlab{}.
\newblock \showarticletitle{Learning phrase representations using RNN
  encoder-decoder for statistical machine translation}.
\newblock \bibinfo{journal}{\emph{arXiv preprint arXiv:1406.1078}}
  (\bibinfo{year}{2014}).
\newblock


\bibitem[\protect\citeauthoryear{Elad and Kimmel}{Elad and Kimmel}{2003}]%
        {elad2003bending}
\bibfield{author}{\bibinfo{person}{Asi Elad} {and} \bibinfo{person}{Ron
  Kimmel}.} \bibinfo{year}{2003}\natexlab{}.
\newblock \showarticletitle{On bending invariant signatures for surfaces}.
\newblock \bibinfo{journal}{\emph{IEEE Transactions on pattern analysis and
  machine intelligence}} \bibinfo{volume}{25}, \bibinfo{number}{10}
  (\bibinfo{year}{2003}), \bibinfo{pages}{1285--1295}.
\newblock


\bibitem[\protect\citeauthoryear{Ezuz, Solomon, Kim, and Ben-Chen}{Ezuz
  et~al\mbox{.}}{2017}]%
        {ezuz2017gwcnn}
\bibfield{author}{\bibinfo{person}{Danielle Ezuz}, \bibinfo{person}{Justin
  Solomon}, \bibinfo{person}{Vladimir~G Kim}, {and} \bibinfo{person}{Mirela
  Ben-Chen}.} \bibinfo{year}{2017}\natexlab{}.
\newblock \showarticletitle{GWCNN: A metric alignment layer for deep shape
  analysis}. In \bibinfo{booktitle}{\emph{Computer Graphics Forum}},
  Vol.~\bibinfo{volume}{36}. Wiley Online Library, \bibinfo{pages}{49--57}.
\newblock


\bibitem[\protect\citeauthoryear{Fanelli, Weise, Gall, and Van~Gool}{Fanelli
  et~al\mbox{.}}{2011}]%
        {fanelli2011real}
\bibfield{author}{\bibinfo{person}{Gabriele Fanelli}, \bibinfo{person}{Thibaut
  Weise}, \bibinfo{person}{Juergen Gall}, {and} \bibinfo{person}{Luc
  Van~Gool}.} \bibinfo{year}{2011}\natexlab{}.
\newblock \showarticletitle{Real time head pose estimation from consumer depth
  cameras}. In \bibinfo{booktitle}{\emph{Joint pattern recognition symposium}}.
  Springer, \bibinfo{pages}{101--110}.
\newblock


\bibitem[\protect\citeauthoryear{Feng, Feng, You, Zhao, and Gao}{Feng
  et~al\mbox{.}}{2019}]%
        {feng2019meshnet}
\bibfield{author}{\bibinfo{person}{Yutong Feng}, \bibinfo{person}{Yifan Feng},
  \bibinfo{person}{Haoxuan You}, \bibinfo{person}{Xibin Zhao}, {and}
  \bibinfo{person}{Yue Gao}.} \bibinfo{year}{2019}\natexlab{}.
\newblock \showarticletitle{MeshNet: mesh neural network for 3D shape
  representation}. In \bibinfo{booktitle}{\emph{Proceedings of the AAAI
  Conference on Artificial Intelligence}}, Vol.~\bibinfo{volume}{33}.
  \bibinfo{pages}{8279--8286}.
\newblock


\bibitem[\protect\citeauthoryear{Feng, Zhang, Zhao, Ji, and Gao}{Feng
  et~al\mbox{.}}{2018a}]%
        {Feng_2018_CVPR}
\bibfield{author}{\bibinfo{person}{Yifan Feng}, \bibinfo{person}{Zizhao Zhang},
  \bibinfo{person}{Xibin Zhao}, \bibinfo{person}{Rongrong Ji}, {and}
  \bibinfo{person}{Yue Gao}.} \bibinfo{year}{2018}\natexlab{a}.
\newblock \showarticletitle{GVCNN: Group-View Convolutional Neural Networks for
  3D Shape Recognition}. In \bibinfo{booktitle}{\emph{The IEEE Conference on
  Computer Vision and Pattern Recognition (CVPR)}}.
\newblock


\bibitem[\protect\citeauthoryear{Feng, Zhang, Zhao, Ji, and Gao}{Feng
  et~al\mbox{.}}{2018b}]%
        {feng2018gvcnn}
\bibfield{author}{\bibinfo{person}{Yifan Feng}, \bibinfo{person}{Zizhao Zhang},
  \bibinfo{person}{Xibin Zhao}, \bibinfo{person}{Rongrong Ji}, {and}
  \bibinfo{person}{Yue Gao}.} \bibinfo{year}{2018}\natexlab{b}.
\newblock \showarticletitle{GVCNN: Group-view convolutional neural networks for
  3D shape recognition}. In \bibinfo{booktitle}{\emph{Proceedings of the IEEE
  Conference on Computer Vision and Pattern Recognition}}.
  \bibinfo{pages}{264--272}.
\newblock


\bibitem[\protect\citeauthoryear{Garland and Heckbert}{Garland and
  Heckbert}{1997}]%
        {garland1997surface}
\bibfield{author}{\bibinfo{person}{Michael Garland} {and}
  \bibinfo{person}{Paul~S Heckbert}.} \bibinfo{year}{1997}\natexlab{}.
\newblock \showarticletitle{Surface simplification using quadric error
  metrics}. In \bibinfo{booktitle}{\emph{Proceedings of the 24th annual
  conference on Computer graphics and interactive techniques}}.
  \bibinfo{pages}{209--216}.
\newblock


\bibitem[\protect\citeauthoryear{Gelfand and Guibas}{Gelfand and
  Guibas}{2004}]%
        {gelfand2004shape}
\bibfield{author}{\bibinfo{person}{Natasha Gelfand} {and}
  \bibinfo{person}{Leonidas~J Guibas}.} \bibinfo{year}{2004}\natexlab{}.
\newblock \showarticletitle{Shape segmentation using local slippage analysis}.
  In \bibinfo{booktitle}{\emph{Proceedings of the 2004 Eurographics/ACM
  SIGGRAPH symposium on Geometry processing}}. \bibinfo{pages}{214--223}.
\newblock


\bibitem[\protect\citeauthoryear{Gezawa, Zhang, Wang, and Yunqi}{Gezawa
  et~al\mbox{.}}{2020}]%
        {gezawa2020review}
\bibfield{author}{\bibinfo{person}{Abubakar~Sulaiman Gezawa},
  \bibinfo{person}{Yan Zhang}, \bibinfo{person}{Qicong Wang}, {and}
  \bibinfo{person}{Lei Yunqi}.} \bibinfo{year}{2020}\natexlab{}.
\newblock \showarticletitle{A Review on Deep Learning Approaches for 3D Data
  Representations in Retrieval and Classifications}.
\newblock \bibinfo{journal}{\emph{IEEE Access}}  \bibinfo{volume}{8}
  (\bibinfo{year}{2020}), \bibinfo{pages}{57566--57593}.
\newblock


\bibitem[\protect\citeauthoryear{Giorgi, Biasotti, and Paraboschi}{Giorgi
  et~al\mbox{.}}{2007}]%
        {giorgi2007shape}
\bibfield{author}{\bibinfo{person}{Daniela Giorgi}, \bibinfo{person}{Silvia
  Biasotti}, {and} \bibinfo{person}{Laura Paraboschi}.}
  \bibinfo{year}{2007}\natexlab{}.
\newblock \showarticletitle{Shape retrieval contest 2007: Watertight models
  track}.
\newblock \bibinfo{journal}{\emph{SHREC competition}} \bibinfo{volume}{8},
  \bibinfo{number}{7} (\bibinfo{year}{2007}).
\newblock


\bibitem[\protect\citeauthoryear{Gomez-Donoso, Garcia-Garcia, Garcia-Rodriguez,
  Orts-Escolano, and Cazorla}{Gomez-Donoso et~al\mbox{.}}{2017}]%
        {gomez2017lonchanet}
\bibfield{author}{\bibinfo{person}{Francisco Gomez-Donoso},
  \bibinfo{person}{Alberto Garcia-Garcia}, \bibinfo{person}{J
  Garcia-Rodriguez}, \bibinfo{person}{Sergio Orts-Escolano}, {and}
  \bibinfo{person}{Miguel Cazorla}.} \bibinfo{year}{2017}\natexlab{}.
\newblock \showarticletitle{Lonchanet: A sliced-based cnn architecture for
  real-time 3d object recognition}. In \bibinfo{booktitle}{\emph{2017
  International Joint Conference on Neural Networks (IJCNN)}}. IEEE,
  \bibinfo{pages}{412--418}.
\newblock


\bibitem[\protect\citeauthoryear{Gong, Chen, Bronstein, and Zafeiriou}{Gong
  et~al\mbox{.}}{2019}]%
        {gong2019spiralnet++}
\bibfield{author}{\bibinfo{person}{Shunwang Gong}, \bibinfo{person}{Lei Chen},
  \bibinfo{person}{Michael Bronstein}, {and} \bibinfo{person}{Stefanos
  Zafeiriou}.} \bibinfo{year}{2019}\natexlab{}.
\newblock \showarticletitle{Spiralnet++: A fast and highly efficient mesh
  convolution operator}. In \bibinfo{booktitle}{\emph{Proceedings of the IEEE
  International Conference on Computer Vision Workshops}}.
  \bibinfo{pages}{0--0}.
\newblock


\bibitem[\protect\citeauthoryear{Gotsman}{Gotsman}{2003}]%
        {gotsman2003graph}
\bibfield{author}{\bibinfo{person}{Craig Gotsman}.}
  \bibinfo{year}{2003}\natexlab{}.
\newblock \showarticletitle{On graph partitioning, spectral analysis, and
  digital mesh processing}. In \bibinfo{booktitle}{\emph{2003 Shape Modeling
  International.}} IEEE, \bibinfo{pages}{165--171}.
\newblock


\bibitem[\protect\citeauthoryear{Grady}{Grady}{2006}]%
        {grady2006random}
\bibfield{author}{\bibinfo{person}{Leo Grady}.}
  \bibinfo{year}{2006}\natexlab{}.
\newblock \showarticletitle{Random walks for image segmentation}.
\newblock \bibinfo{journal}{\emph{IEEE transactions on pattern analysis and
  machine intelligence}} \bibinfo{volume}{28}, \bibinfo{number}{11}
  (\bibinfo{year}{2006}), \bibinfo{pages}{1768--1783}.
\newblock


\bibitem[\protect\citeauthoryear{Graves, Liwicki, Fern{\'a}ndez, Bertolami,
  Bunke, and Schmidhuber}{Graves et~al\mbox{.}}{2008}]%
        {graves2008novel}
\bibfield{author}{\bibinfo{person}{Alex Graves}, \bibinfo{person}{Marcus
  Liwicki}, \bibinfo{person}{Santiago Fern{\'a}ndez}, \bibinfo{person}{Roman
  Bertolami}, \bibinfo{person}{Horst Bunke}, {and} \bibinfo{person}{J{\"u}rgen
  Schmidhuber}.} \bibinfo{year}{2008}\natexlab{}.
\newblock \showarticletitle{A novel connectionist system for unconstrained
  handwriting recognition}.
\newblock \bibinfo{journal}{\emph{IEEE transactions on pattern analysis and
  machine intelligence}} \bibinfo{volume}{31}, \bibinfo{number}{5}
  (\bibinfo{year}{2008}), \bibinfo{pages}{855--868}.
\newblock


\bibitem[\protect\citeauthoryear{Guerrero, Kleiman, Ovsjanikov, and
  Mitra}{Guerrero et~al\mbox{.}}{2018}]%
        {guerrero2018pcpnet}
\bibfield{author}{\bibinfo{person}{Paul Guerrero}, \bibinfo{person}{Yanir
  Kleiman}, \bibinfo{person}{Maks Ovsjanikov}, {and} \bibinfo{person}{Niloy~J
  Mitra}.} \bibinfo{year}{2018}\natexlab{}.
\newblock \showarticletitle{PCPNet learning local shape properties from raw
  point clouds}. In \bibinfo{booktitle}{\emph{Computer Graphics Forum}},
  Vol.~\bibinfo{volume}{37}. Wiley Online Library, \bibinfo{pages}{75--85}.
\newblock


\bibitem[\protect\citeauthoryear{Guo, Zou, and Chen}{Guo et~al\mbox{.}}{2015}]%
        {guo20153d}
\bibfield{author}{\bibinfo{person}{Kan Guo}, \bibinfo{person}{Dongqing Zou},
  {and} \bibinfo{person}{Xiaowu Chen}.} \bibinfo{year}{2015}\natexlab{}.
\newblock \showarticletitle{3d mesh labeling via deep convolutional neural
  networks}.
\newblock \bibinfo{journal}{\emph{ACM Transactions on Graphics (TOG)}}
  \bibinfo{volume}{35}, \bibinfo{number}{1} (\bibinfo{year}{2015}),
  \bibinfo{pages}{1--12}.
\newblock


\bibitem[\protect\citeauthoryear{Haim, Segol, Ben-Hamu, Maron, and Lipman}{Haim
  et~al\mbox{.}}{2019}]%
        {haim2019surface}
\bibfield{author}{\bibinfo{person}{Niv Haim}, \bibinfo{person}{Nimrod Segol},
  \bibinfo{person}{Heli Ben-Hamu}, \bibinfo{person}{Haggai Maron}, {and}
  \bibinfo{person}{Yaron Lipman}.} \bibinfo{year}{2019}\natexlab{}.
\newblock \showarticletitle{Surface Networks via General Covers}. In
  \bibinfo{booktitle}{\emph{Proceedings of the IEEE International Conference on
  Computer Vision}}. \bibinfo{pages}{632--641}.
\newblock


\bibitem[\protect\citeauthoryear{Han, Lu, Liu, Vong, Liu, Zwicker, Han, and
  Chen}{Han et~al\mbox{.}}{2019}]%
        {han20193d2seqviews}
\bibfield{author}{\bibinfo{person}{Zhizhong Han}, \bibinfo{person}{Honglei Lu},
  \bibinfo{person}{Zhenbao Liu}, \bibinfo{person}{Chi-Man Vong},
  \bibinfo{person}{Yu-Shen Liu}, \bibinfo{person}{Matthias Zwicker},
  \bibinfo{person}{Junwei Han}, {and} \bibinfo{person}{CL~Philip Chen}.}
  \bibinfo{year}{2019}\natexlab{}.
\newblock \showarticletitle{3d2seqviews: Aggregating sequential views for 3d
  global feature learning by cnn with hierarchical attention aggregation}.
\newblock \bibinfo{journal}{\emph{IEEE Transactions on Image Processing}}
  \bibinfo{volume}{28}, \bibinfo{number}{8} (\bibinfo{year}{2019}),
  \bibinfo{pages}{3986--3999}.
\newblock


\bibitem[\protect\citeauthoryear{Hanocka, Fish, Wang, Giryes, Fleishman, and
  Cohen-Or}{Hanocka et~al\mbox{.}}{2018}]%
        {hanocka2018alignet}
\bibfield{author}{\bibinfo{person}{Rana Hanocka}, \bibinfo{person}{Noa Fish},
  \bibinfo{person}{Zhenhua Wang}, \bibinfo{person}{Raja Giryes},
  \bibinfo{person}{Shachar Fleishman}, {and} \bibinfo{person}{Daniel
  Cohen-Or}.} \bibinfo{year}{2018}\natexlab{}.
\newblock \showarticletitle{Alignet: Partial-shape agnostic alignment via
  unsupervised learning}.
\newblock \bibinfo{journal}{\emph{ACM Transactions on Graphics (TOG)}}
  \bibinfo{volume}{38}, \bibinfo{number}{1} (\bibinfo{year}{2018}),
  \bibinfo{pages}{1--14}.
\newblock


\bibitem[\protect\citeauthoryear{Hanocka, Hertz, Fish, Giryes, Fleishman, and
  Cohen-Or}{Hanocka et~al\mbox{.}}{2019}]%
        {hanocka2019meshcnn}
\bibfield{author}{\bibinfo{person}{Rana Hanocka}, \bibinfo{person}{Amir Hertz},
  \bibinfo{person}{Noa Fish}, \bibinfo{person}{Raja Giryes},
  \bibinfo{person}{Shachar Fleishman}, {and} \bibinfo{person}{Daniel
  Cohen-Or}.} \bibinfo{year}{2019}\natexlab{}.
\newblock \showarticletitle{MeshCNN: a network with an edge}.
\newblock \bibinfo{journal}{\emph{ACM Transactions on Graphics (TOG)}}
  \bibinfo{volume}{38}, \bibinfo{number}{4} (\bibinfo{year}{2019}),
  \bibinfo{pages}{1--12}.
\newblock


\bibitem[\protect\citeauthoryear{He, Zhou, Zhou, Bai, and Bai}{He
  et~al\mbox{.}}{2018}]%
        {he2018triplet}
\bibfield{author}{\bibinfo{person}{Xinwei He}, \bibinfo{person}{Yang Zhou},
  \bibinfo{person}{Zhichao Zhou}, \bibinfo{person}{Song Bai}, {and}
  \bibinfo{person}{Xiang Bai}.} \bibinfo{year}{2018}\natexlab{}.
\newblock \showarticletitle{Triplet-center loss for multi-view 3d object
  retrieval}. In \bibinfo{booktitle}{\emph{Proceedings of the IEEE Conference
  on Computer Vision and Pattern Recognition}}. \bibinfo{pages}{1945--1954}.
\newblock


\bibitem[\protect\citeauthoryear{Henaff, Bruna, and LeCun}{Henaff
  et~al\mbox{.}}{2015}]%
        {henaff2015deep}
\bibfield{author}{\bibinfo{person}{Mikael Henaff}, \bibinfo{person}{Joan
  Bruna}, {and} \bibinfo{person}{Yann LeCun}.} \bibinfo{year}{2015}\natexlab{}.
\newblock \showarticletitle{Deep convolutional networks on graph-structured
  data}.
\newblock \bibinfo{journal}{\emph{arXiv preprint arXiv:1506.05163}}
  (\bibinfo{year}{2015}).
\newblock


\bibitem[\protect\citeauthoryear{Hilaga, Shinagawa, Kohmura, and Kunii}{Hilaga
  et~al\mbox{.}}{2001}]%
        {hilaga2001topology}
\bibfield{author}{\bibinfo{person}{Masaki Hilaga}, \bibinfo{person}{Yoshihisa
  Shinagawa}, \bibinfo{person}{Taku Kohmura}, {and} \bibinfo{person}{Tosiyasu~L
  Kunii}.} \bibinfo{year}{2001}\natexlab{}.
\newblock \showarticletitle{Topology matching for fully automatic similarity
  estimation of 3D shapes}. In \bibinfo{booktitle}{\emph{Proceedings of the
  28th annual conference on Computer graphics and interactive techniques}}.
  \bibinfo{pages}{203--212}.
\newblock


\bibitem[\protect\citeauthoryear{Hochreiter and Schmidhuber}{Hochreiter and
  Schmidhuber}{1997}]%
        {hochreiter1997long}
\bibfield{author}{\bibinfo{person}{Sepp Hochreiter} {and}
  \bibinfo{person}{J{\"u}rgen Schmidhuber}.} \bibinfo{year}{1997}\natexlab{}.
\newblock \showarticletitle{Long short-term memory}.
\newblock \bibinfo{journal}{\emph{Neural computation}} \bibinfo{volume}{9},
  \bibinfo{number}{8} (\bibinfo{year}{1997}), \bibinfo{pages}{1735--1780}.
\newblock


\bibitem[\protect\citeauthoryear{Hoppe}{Hoppe}{1997}]%
        {hoppe1997view}
\bibfield{author}{\bibinfo{person}{Hugues Hoppe}.}
  \bibinfo{year}{1997}\natexlab{}.
\newblock \showarticletitle{View-dependent refinement of progressive meshes}.
  In \bibinfo{booktitle}{\emph{Proceedings of the 24th annual conference on
  Computer graphics and interactive techniques}}. \bibinfo{pages}{189--198}.
\newblock


\bibitem[\protect\citeauthoryear{Hua, Tran, and Yeung}{Hua
  et~al\mbox{.}}{2018}]%
        {hua2018pointwise}
\bibfield{author}{\bibinfo{person}{Binh-Son Hua}, \bibinfo{person}{Minh-Khoi
  Tran}, {and} \bibinfo{person}{Sai-Kit Yeung}.}
  \bibinfo{year}{2018}\natexlab{}.
\newblock \showarticletitle{Pointwise convolutional neural networks}. In
  \bibinfo{booktitle}{\emph{Proceedings of the IEEE Conference on Computer
  Vision and Pattern Recognition}}. \bibinfo{pages}{984--993}.
\newblock


\bibitem[\protect\citeauthoryear{Jain and Zhang}{Jain and Zhang}{2007}]%
        {jain2007spectral}
\bibfield{author}{\bibinfo{person}{Varun Jain} {and} \bibinfo{person}{Hao
  Zhang}.} \bibinfo{year}{2007}\natexlab{}.
\newblock \showarticletitle{A spectral approach to shape-based retrieval of
  articulated 3D models}.
\newblock \bibinfo{journal}{\emph{Computer-Aided Design}} \bibinfo{volume}{39},
  \bibinfo{number}{5} (\bibinfo{year}{2007}), \bibinfo{pages}{398--407}.
\newblock


\bibitem[\protect\citeauthoryear{Johns, Leutenegger, and Davison}{Johns
  et~al\mbox{.}}{2016}]%
        {johns2016pairwise}
\bibfield{author}{\bibinfo{person}{Edward Johns}, \bibinfo{person}{Stefan
  Leutenegger}, {and} \bibinfo{person}{Andrew~J Davison}.}
  \bibinfo{year}{2016}\natexlab{}.
\newblock \showarticletitle{Pairwise decomposition of image sequences for
  active multi-view recognition}. In \bibinfo{booktitle}{\emph{Proceedings of
  the IEEE Conference on Computer Vision and Pattern Recognition}}.
  \bibinfo{pages}{3813--3822}.
\newblock


\bibitem[\protect\citeauthoryear{Johnson and Hebert}{Johnson and
  Hebert}{1999}]%
        {johnson1999using}
\bibfield{author}{\bibinfo{person}{Andrew~E. Johnson} {and}
  \bibinfo{person}{Martial Hebert}.} \bibinfo{year}{1999}\natexlab{}.
\newblock \showarticletitle{Using spin images for efficient object recognition
  in cluttered 3D scenes}.
\newblock \bibinfo{journal}{\emph{IEEE Transactions on pattern analysis and
  machine intelligence}} \bibinfo{volume}{21}, \bibinfo{number}{5}
  (\bibinfo{year}{1999}), \bibinfo{pages}{433--449}.
\newblock


\bibitem[\protect\citeauthoryear{Kalogerakis, Averkiou, Maji, and
  Chaudhuri}{Kalogerakis et~al\mbox{.}}{2017}]%
        {kalogerakis20173d}
\bibfield{author}{\bibinfo{person}{Evangelos Kalogerakis},
  \bibinfo{person}{Melinos Averkiou}, \bibinfo{person}{Subhransu Maji}, {and}
  \bibinfo{person}{Siddhartha Chaudhuri}.} \bibinfo{year}{2017}\natexlab{}.
\newblock \showarticletitle{3D shape segmentation with projective convolutional
  networks}. In \bibinfo{booktitle}{\emph{Proceedings of the IEEE Conference on
  Computer Vision and Pattern Recognition}}. \bibinfo{pages}{3779--3788}.
\newblock


\bibitem[\protect\citeauthoryear{Kalogerakis, Hertzmann, and Singh}{Kalogerakis
  et~al\mbox{.}}{2010}]%
        {kalogerakis2010learning}
\bibfield{author}{\bibinfo{person}{Evangelos Kalogerakis},
  \bibinfo{person}{Aaron Hertzmann}, {and} \bibinfo{person}{Karan Singh}.}
  \bibinfo{year}{2010}\natexlab{}.
\newblock \showarticletitle{Learning 3D mesh segmentation and labeling}.
\newblock In \bibinfo{booktitle}{\emph{ACM SIGGRAPH 2010 papers}}.
  \bibinfo{pages}{1--12}.
\newblock


\bibitem[\protect\citeauthoryear{Kanezaki, Matsushita, and Nishida}{Kanezaki
  et~al\mbox{.}}{2018}]%
        {kanezaki2018rotationnet}
\bibfield{author}{\bibinfo{person}{Asako Kanezaki}, \bibinfo{person}{Yasuyuki
  Matsushita}, {and} \bibinfo{person}{Yoshifumi Nishida}.}
  \bibinfo{year}{2018}\natexlab{}.
\newblock \showarticletitle{Rotationnet: Joint object categorization and pose
  estimation using multiviews from unsupervised viewpoints}. In
  \bibinfo{booktitle}{\emph{Proceedings of the IEEE Conference on Computer
  Vision and Pattern Recognition}}. \bibinfo{pages}{5010--5019}.
\newblock


\bibitem[\protect\citeauthoryear{Katz, Leifman, and Tal}{Katz
  et~al\mbox{.}}{2005}]%
        {katz2005mesh}
\bibfield{author}{\bibinfo{person}{Sagi Katz}, \bibinfo{person}{George
  Leifman}, {and} \bibinfo{person}{Ayellet Tal}.}
  \bibinfo{year}{2005}\natexlab{}.
\newblock \showarticletitle{Mesh segmentation using feature point and core
  extraction}.
\newblock \bibinfo{journal}{\emph{The Visual Computer}} \bibinfo{volume}{21},
  \bibinfo{number}{8-10} (\bibinfo{year}{2005}), \bibinfo{pages}{649--658}.
\newblock


\bibitem[\protect\citeauthoryear{Katz and Tal}{Katz and Tal}{2003}]%
        {katz2003hierarchical}
\bibfield{author}{\bibinfo{person}{Sagi Katz} {and} \bibinfo{person}{Ayellet
  Tal}.} \bibinfo{year}{2003}\natexlab{}.
\newblock \showarticletitle{Hierarchical mesh decomposition using fuzzy
  clustering and cuts}.
\newblock \bibinfo{journal}{\emph{ACM transactions on graphics (TOG)}}
  \bibinfo{volume}{22}, \bibinfo{number}{3} (\bibinfo{year}{2003}),
  \bibinfo{pages}{954--961}.
\newblock


\bibitem[\protect\citeauthoryear{Kingma and Ba}{Kingma and Ba}{2014}]%
        {kingma2014adam}
\bibfield{author}{\bibinfo{person}{Diederik~P Kingma} {and}
  \bibinfo{person}{Jimmy Ba}.} \bibinfo{year}{2014}\natexlab{}.
\newblock \showarticletitle{Adam: A method for stochastic optimization}.
\newblock \bibinfo{journal}{\emph{arXiv preprint arXiv:1412.6980}}
  (\bibinfo{year}{2014}).
\newblock


\bibitem[\protect\citeauthoryear{Kipf and Welling}{Kipf and Welling}{2016}]%
        {kipf2016semi}
\bibfield{author}{\bibinfo{person}{Thomas~N Kipf} {and} \bibinfo{person}{Max
  Welling}.} \bibinfo{year}{2016}\natexlab{}.
\newblock \showarticletitle{Semi-supervised classification with graph
  convolutional networks}.
\newblock \bibinfo{journal}{\emph{arXiv preprint arXiv:1609.02907}}
  (\bibinfo{year}{2016}).
\newblock


\bibitem[\protect\citeauthoryear{Koschan}{Koschan}{2003}]%
        {koschan2003perception}
\bibfield{author}{\bibinfo{person}{AF Koschan}.}
  \bibinfo{year}{2003}\natexlab{}.
\newblock \showarticletitle{Perception-based 3D triangle mesh segmentation
  using fast marching watersheds}. In \bibinfo{booktitle}{\emph{2003 IEEE
  Computer Society Conference on Computer Vision and Pattern Recognition, 2003.
  Proceedings.}}, Vol.~\bibinfo{volume}{2}. IEEE, \bibinfo{pages}{II--II}.
\newblock


\bibitem[\protect\citeauthoryear{Lai, Hu, Martin, and Rosin}{Lai
  et~al\mbox{.}}{2008}]%
        {Lai:2008:FMS:1364901.1364927}
\bibfield{author}{\bibinfo{person}{Yu-Kun Lai}, \bibinfo{person}{Shi-Min Hu},
  \bibinfo{person}{Ralph~R. Martin}, {and} \bibinfo{person}{Paul~L. Rosin}.}
  \bibinfo{year}{2008}\natexlab{}.
\newblock \showarticletitle{Fast Mesh Segmentation Using Random Walks}. In
  \bibinfo{booktitle}{\emph{Proceedings of the 2008 ACM Symposium on Solid and
  Physical Modeling}} \emph{(\bibinfo{series}{SPM '08})}.
  \bibinfo{publisher}{ACM}, \bibinfo{address}{New York, NY, USA},
  \bibinfo{pages}{183--191}.
\newblock
\showISBNx{978-1-60558-106-4}
\urldef\tempurl%
\url{https://doi.org/10.1145/1364901.1364927}
\showDOI{\tempurl}


\bibitem[\protect\citeauthoryear{Lavou{\'e}, Dupont, and Baskurt}{Lavou{\'e}
  et~al\mbox{.}}{2005}]%
        {lavoue2005new}
\bibfield{author}{\bibinfo{person}{Guillaume Lavou{\'e}},
  \bibinfo{person}{Florent Dupont}, {and} \bibinfo{person}{Atilla Baskurt}.}
  \bibinfo{year}{2005}\natexlab{}.
\newblock \showarticletitle{A new CAD mesh segmentation method, based on
  curvature tensor analysis}.
\newblock \bibinfo{journal}{\emph{Computer-Aided Design}} \bibinfo{volume}{37},
  \bibinfo{number}{10} (\bibinfo{year}{2005}), \bibinfo{pages}{975--987}.
\newblock


\bibitem[\protect\citeauthoryear{Li, Bu, Sun, Wu, Di, and Chen}{Li
  et~al\mbox{.}}{2018}]%
        {li2018pointcnn}
\bibfield{author}{\bibinfo{person}{Yangyan Li}, \bibinfo{person}{Rui Bu},
  \bibinfo{person}{Mingchao Sun}, \bibinfo{person}{Wei Wu},
  \bibinfo{person}{Xinhan Di}, {and} \bibinfo{person}{Baoquan Chen}.}
  \bibinfo{year}{2018}\natexlab{}.
\newblock \showarticletitle{Pointcnn: Convolution on x-transformed points}. In
  \bibinfo{booktitle}{\emph{Advances in neural information processing
  systems}}. \bibinfo{pages}{820--830}.
\newblock


\bibitem[\protect\citeauthoryear{Lian, Godil, Bustos, Daoudi, Hermans,
  Kawamura, Kurita, Lavoua, and Dp~Suetens}{Lian et~al\mbox{.}}{2011}]%
        {lian2011shape}
\bibfield{author}{\bibinfo{person}{Z Lian}, \bibinfo{person}{A Godil},
  \bibinfo{person}{B Bustos}, \bibinfo{person}{M Daoudi}, \bibinfo{person}{J
  Hermans}, \bibinfo{person}{S Kawamura}, \bibinfo{person}{Y Kurita},
  \bibinfo{person}{G Lavoua}, {and} \bibinfo{person}{P Dp~Suetens}.}
  \bibinfo{year}{2011}\natexlab{}.
\newblock \showarticletitle{Shape retrieval on non-rigid 3D watertight meshes}.
  In \bibinfo{booktitle}{\emph{Eurographics workshop on 3d object retrieval
  (3DOR)}}. Citeseer.
\newblock


\bibitem[\protect\citeauthoryear{Lian, Godil, Bustos, Daoudi, Hermans,
  Kawamura, Kurita, Lavou{\'e}, Van~Nguyen, Ohbuchi, et~al\mbox{.}}{Lian
  et~al\mbox{.}}{2013}]%
        {lian2013comparison}
\bibfield{author}{\bibinfo{person}{Zhouhui Lian}, \bibinfo{person}{Afzal
  Godil}, \bibinfo{person}{Benjamin Bustos}, \bibinfo{person}{Mohamed Daoudi},
  \bibinfo{person}{Jeroen Hermans}, \bibinfo{person}{Shun Kawamura},
  \bibinfo{person}{Yukinori Kurita}, \bibinfo{person}{Guillaume Lavou{\'e}},
  \bibinfo{person}{Hien Van~Nguyen}, \bibinfo{person}{Ryutarou Ohbuchi},
  {et~al\mbox{.}}} \bibinfo{year}{2013}\natexlab{}.
\newblock \showarticletitle{A comparison of methods for non-rigid 3D shape
  retrieval}.
\newblock \bibinfo{journal}{\emph{Pattern Recognition}} \bibinfo{volume}{46},
  \bibinfo{number}{1} (\bibinfo{year}{2013}), \bibinfo{pages}{449--461}.
\newblock


\bibitem[\protect\citeauthoryear{Lim, Dielen, Campen, and Kobbelt}{Lim
  et~al\mbox{.}}{2018}]%
        {lim2018simple}
\bibfield{author}{\bibinfo{person}{Isaak Lim}, \bibinfo{person}{Alexander
  Dielen}, \bibinfo{person}{Marcel Campen}, {and} \bibinfo{person}{Leif
  Kobbelt}.} \bibinfo{year}{2018}\natexlab{}.
\newblock \showarticletitle{A simple approach to intrinsic correspondence
  learning on unstructured 3d meshes}. In \bibinfo{booktitle}{\emph{Proceedings
  of the European Conference on Computer Vision (ECCV)}}.
  \bibinfo{pages}{0--0}.
\newblock


\bibitem[\protect\citeauthoryear{Liu and Zhang}{Liu and Zhang}{2004}]%
        {liu2004segmentation}
\bibfield{author}{\bibinfo{person}{Rong Liu} {and} \bibinfo{person}{Hao
  Zhang}.} \bibinfo{year}{2004}\natexlab{}.
\newblock \showarticletitle{Segmentation of 3D meshes through spectral
  clustering}. In \bibinfo{booktitle}{\emph{12th Pacific Conference on Computer
  Graphics and Applications, 2004. PG 2004. Proceedings.}} IEEE,
  \bibinfo{pages}{298--305}.
\newblock


\bibitem[\protect\citeauthoryear{Liu, Fan, Xiang, and Pan}{Liu
  et~al\mbox{.}}{2019}]%
        {liu2019relation}
\bibfield{author}{\bibinfo{person}{Yongcheng Liu}, \bibinfo{person}{Bin Fan},
  \bibinfo{person}{Shiming Xiang}, {and} \bibinfo{person}{Chunhong Pan}.}
  \bibinfo{year}{2019}\natexlab{}.
\newblock \showarticletitle{Relation-shape convolutional neural network for
  point cloud analysis}. In \bibinfo{booktitle}{\emph{Proceedings of the IEEE
  Conference on Computer Vision and Pattern Recognition}}.
  \bibinfo{pages}{8895--8904}.
\newblock


\bibitem[\protect\citeauthoryear{Liu, Zha, and Qin}{Liu et~al\mbox{.}}{2006}]%
        {liu2006shape}
\bibfield{author}{\bibinfo{person}{Yi Liu}, \bibinfo{person}{Hongbin Zha},
  {and} \bibinfo{person}{Hong Qin}.} \bibinfo{year}{2006}\natexlab{}.
\newblock \showarticletitle{Shape topics: A compact representation and new
  algorithms for 3d partial shape retrieval}. In \bibinfo{booktitle}{\emph{2006
  IEEE Computer Society Conference on Computer Vision and Pattern Recognition
  (CVPR'06)}}, Vol.~\bibinfo{volume}{2}. IEEE, \bibinfo{pages}{2025--2032}.
\newblock


\bibitem[\protect\citeauthoryear{Lov{\'a}sz et~al\mbox{.}}{Lov{\'a}sz
  et~al\mbox{.}}{1993}]%
        {lovasz1993random}
\bibfield{author}{\bibinfo{person}{L{\'a}szl{\'o} Lov{\'a}sz} {et~al\mbox{.}}}
  \bibinfo{year}{1993}\natexlab{}.
\newblock \showarticletitle{Random walks on graphs: A survey}.
\newblock \bibinfo{journal}{\emph{Combinatorics, Paul erdos is eighty}}
  \bibinfo{volume}{2}, \bibinfo{number}{1} (\bibinfo{year}{1993}),
  \bibinfo{pages}{1--46}.
\newblock


\bibitem[\protect\citeauthoryear{Lowe}{Lowe}{2004}]%
        {lowe2004distinctive}
\bibfield{author}{\bibinfo{person}{David~G Lowe}.}
  \bibinfo{year}{2004}\natexlab{}.
\newblock \showarticletitle{Distinctive image features from scale-invariant
  keypoints}.
\newblock \bibinfo{journal}{\emph{International journal of computer vision}}
  \bibinfo{volume}{60}, \bibinfo{number}{2} (\bibinfo{year}{2004}),
  \bibinfo{pages}{91--110}.
\newblock


\bibitem[\protect\citeauthoryear{Mahmoudi and Sapiro}{Mahmoudi and
  Sapiro}{2009}]%
        {mahmoudi2009three}
\bibfield{author}{\bibinfo{person}{Mona Mahmoudi} {and}
  \bibinfo{person}{Guillermo Sapiro}.} \bibinfo{year}{2009}\natexlab{}.
\newblock \showarticletitle{Three-dimensional point cloud recognition via
  distributions of geometric distances}.
\newblock \bibinfo{journal}{\emph{Graphical Models}} \bibinfo{volume}{71},
  \bibinfo{number}{1} (\bibinfo{year}{2009}), \bibinfo{pages}{22--31}.
\newblock


\bibitem[\protect\citeauthoryear{Maron, Galun, Aigerman, Trope, Dym, Yumer,
  Kim, and Lipman}{Maron et~al\mbox{.}}{2017}]%
        {maron2017convolutional}
\bibfield{author}{\bibinfo{person}{Haggai Maron}, \bibinfo{person}{Meirav
  Galun}, \bibinfo{person}{Noam Aigerman}, \bibinfo{person}{Miri Trope},
  \bibinfo{person}{Nadav Dym}, \bibinfo{person}{Ersin Yumer},
  \bibinfo{person}{Vladimir~G Kim}, {and} \bibinfo{person}{Yaron Lipman}.}
  \bibinfo{year}{2017}\natexlab{}.
\newblock \showarticletitle{Convolutional neural networks on surfaces via
  seamless toric covers.}
\newblock \bibinfo{journal}{\emph{ACM Trans. Graph.}} \bibinfo{volume}{36},
  \bibinfo{number}{4} (\bibinfo{year}{2017}), \bibinfo{pages}{71--1}.
\newblock


\bibitem[\protect\citeauthoryear{Masci, Boscaini, Bronstein, and
  Vandergheynst}{Masci et~al\mbox{.}}{2015}]%
        {masci2015geodesic}
\bibfield{author}{\bibinfo{person}{Jonathan Masci}, \bibinfo{person}{Davide
  Boscaini}, \bibinfo{person}{Michael Bronstein}, {and} \bibinfo{person}{Pierre
  Vandergheynst}.} \bibinfo{year}{2015}\natexlab{}.
\newblock \showarticletitle{Geodesic convolutional neural networks on
  riemannian manifolds}. In \bibinfo{booktitle}{\emph{Proceedings of the IEEE
  international conference on computer vision workshops}}.
  \bibinfo{pages}{37--45}.
\newblock


\bibitem[\protect\citeauthoryear{Maturana and Scherer}{Maturana and
  Scherer}{2015}]%
        {maturana2015voxnet}
\bibfield{author}{\bibinfo{person}{Daniel Maturana} {and}
  \bibinfo{person}{Sebastian Scherer}.} \bibinfo{year}{2015}\natexlab{}.
\newblock \showarticletitle{Voxnet: A 3d convolutional neural network for
  real-time object recognition}. In \bibinfo{booktitle}{\emph{2015 IEEE/RSJ
  International Conference on Intelligent Robots and Systems (IROS)}}. IEEE,
  \bibinfo{pages}{922--928}.
\newblock


\bibitem[\protect\citeauthoryear{M{\'e}moli}{M{\'e}moli}{2007}]%
        {memoli2007use}
\bibfield{author}{\bibinfo{person}{Facundo M{\'e}moli}.}
  \bibinfo{year}{2007}\natexlab{}.
\newblock \showarticletitle{On the use of Gromov-Hausdorff distances for shape
  comparison}.
\newblock  (\bibinfo{year}{2007}).
\newblock


\bibitem[\protect\citeauthoryear{M{\'e}moli and Sapiro}{M{\'e}moli and
  Sapiro}{2005}]%
        {memoli2005theoretical}
\bibfield{author}{\bibinfo{person}{Facundo M{\'e}moli} {and}
  \bibinfo{person}{Guillermo Sapiro}.} \bibinfo{year}{2005}\natexlab{}.
\newblock \showarticletitle{A theoretical and computational framework for
  isometry invariant recognition of point cloud data}.
\newblock \bibinfo{journal}{\emph{Foundations of Computational Mathematics}}
  \bibinfo{volume}{5}, \bibinfo{number}{3} (\bibinfo{year}{2005}),
  \bibinfo{pages}{313--347}.
\newblock


\bibitem[\protect\citeauthoryear{Noh and Rieger}{Noh and Rieger}{2004}]%
        {noh2004random}
\bibfield{author}{\bibinfo{person}{Jae~Dong Noh} {and} \bibinfo{person}{Heiko
  Rieger}.} \bibinfo{year}{2004}\natexlab{}.
\newblock \showarticletitle{Random walks on complex networks}.
\newblock \bibinfo{journal}{\emph{Physical review letters}}
  \bibinfo{volume}{92}, \bibinfo{number}{11} (\bibinfo{year}{2004}),
  \bibinfo{pages}{118701}.
\newblock


\bibitem[\protect\citeauthoryear{Ovsjanikov, Bronstein, Bronstein, and
  Guibas}{Ovsjanikov et~al\mbox{.}}{2009}]%
        {ovsjanikov2009shape}
\bibfield{author}{\bibinfo{person}{Maks Ovsjanikov},
  \bibinfo{person}{Alexander~M Bronstein}, \bibinfo{person}{Michael~M
  Bronstein}, {and} \bibinfo{person}{Leonidas~J Guibas}.}
  \bibinfo{year}{2009}\natexlab{}.
\newblock \showarticletitle{Shape google: a computer vision approach to
  isometry invariant shape retrieval}. In \bibinfo{booktitle}{\emph{2009 IEEE
  12th International Conference on Computer Vision Workshops, ICCV Workshops}}.
  IEEE, \bibinfo{pages}{320--327}.
\newblock


\bibitem[\protect\citeauthoryear{Perozzi, Al-Rfou, and Skiena}{Perozzi
  et~al\mbox{.}}{2014}]%
        {perozzi2014deepwalk}
\bibfield{author}{\bibinfo{person}{Bryan Perozzi}, \bibinfo{person}{Rami
  Al-Rfou}, {and} \bibinfo{person}{Steven Skiena}.}
  \bibinfo{year}{2014}\natexlab{}.
\newblock \showarticletitle{Deepwalk: Online learning of social
  representations}. In \bibinfo{booktitle}{\emph{Proceedings of the 20th ACM
  SIGKDD international conference on Knowledge discovery and data mining}}.
  \bibinfo{pages}{701--710}.
\newblock


\bibitem[\protect\citeauthoryear{Poulenard and Ovsjanikov}{Poulenard and
  Ovsjanikov}{2018}]%
        {poulenard2018multi}
\bibfield{author}{\bibinfo{person}{Adrien Poulenard} {and}
  \bibinfo{person}{Maks Ovsjanikov}.} \bibinfo{year}{2018}\natexlab{}.
\newblock \showarticletitle{Multi-directional geodesic neural networks via
  equivariant convolution}.
\newblock \bibinfo{journal}{\emph{ACM Transactions on Graphics (TOG)}}
  \bibinfo{volume}{37}, \bibinfo{number}{6} (\bibinfo{year}{2018}),
  \bibinfo{pages}{1--14}.
\newblock


\bibitem[\protect\citeauthoryear{Qi, Su, Mo, and Guibas}{Qi
  et~al\mbox{.}}{2017a}]%
        {qi2017pointnet}
\bibfield{author}{\bibinfo{person}{Charles~R Qi}, \bibinfo{person}{Hao Su},
  \bibinfo{person}{Kaichun Mo}, {and} \bibinfo{person}{Leonidas~J Guibas}.}
  \bibinfo{year}{2017}\natexlab{a}.
\newblock \showarticletitle{Pointnet: Deep learning on point sets for 3d
  classification and segmentation}. In \bibinfo{booktitle}{\emph{Proceedings of
  the IEEE conference on computer vision and pattern recognition}}.
  \bibinfo{pages}{652--660}.
\newblock


\bibitem[\protect\citeauthoryear{Qi, Su, Nie{\ss}ner, Dai, Yan, and Guibas}{Qi
  et~al\mbox{.}}{2016}]%
        {qi2016volumetric}
\bibfield{author}{\bibinfo{person}{Charles~R Qi}, \bibinfo{person}{Hao Su},
  \bibinfo{person}{Matthias Nie{\ss}ner}, \bibinfo{person}{Angela Dai},
  \bibinfo{person}{Mengyuan Yan}, {and} \bibinfo{person}{Leonidas~J Guibas}.}
  \bibinfo{year}{2016}\natexlab{}.
\newblock \showarticletitle{Volumetric and multi-view cnns for object
  classification on 3d data}. In \bibinfo{booktitle}{\emph{Proceedings of the
  IEEE conference on computer vision and pattern recognition}}.
  \bibinfo{pages}{5648--5656}.
\newblock


\bibitem[\protect\citeauthoryear{Qi, Yi, Su, and Guibas}{Qi
  et~al\mbox{.}}{2017b}]%
        {qi2017pointnet++}
\bibfield{author}{\bibinfo{person}{Charles~Ruizhongtai Qi}, \bibinfo{person}{Li
  Yi}, \bibinfo{person}{Hao Su}, {and} \bibinfo{person}{Leonidas~J Guibas}.}
  \bibinfo{year}{2017}\natexlab{b}.
\newblock \showarticletitle{Pointnet++: Deep hierarchical feature learning on
  point sets in a metric space}. In \bibinfo{booktitle}{\emph{Advances in
  neural information processing systems}}. \bibinfo{pages}{5099--5108}.
\newblock


\bibitem[\protect\citeauthoryear{Reuter, Wolter, and Peinecke}{Reuter
  et~al\mbox{.}}{2005}]%
        {reuter2005laplace}
\bibfield{author}{\bibinfo{person}{Martin Reuter}, \bibinfo{person}{Franz-Erich
  Wolter}, {and} \bibinfo{person}{Niklas Peinecke}.}
  \bibinfo{year}{2005}\natexlab{}.
\newblock \showarticletitle{Laplace-spectra as fingerprints for shape
  matching}. In \bibinfo{booktitle}{\emph{Proceedings of the 2005 ACM symposium
  on Solid and physical modeling}}. \bibinfo{pages}{101--106}.
\newblock


\bibitem[\protect\citeauthoryear{Rodrigues, Morgado, and Gomes}{Rodrigues
  et~al\mbox{.}}{2018}]%
        {rodrigues2018part}
\bibfield{author}{\bibinfo{person}{Rui~SV Rodrigues},
  \bibinfo{person}{Jos{\'e}~FM Morgado}, {and} \bibinfo{person}{Abel~JP
  Gomes}.} \bibinfo{year}{2018}\natexlab{}.
\newblock \showarticletitle{Part-based mesh segmentation: a survey}. In
  \bibinfo{booktitle}{\emph{Computer Graphics Forum}},
  Vol.~\bibinfo{volume}{37}. Wiley Online Library, \bibinfo{pages}{235--274}.
\newblock


\bibitem[\protect\citeauthoryear{Roynard, Deschaud, and Goulette}{Roynard
  et~al\mbox{.}}{2018}]%
        {roynard2018classification}
\bibfield{author}{\bibinfo{person}{Xavier Roynard},
  \bibinfo{person}{Jean-Emmanuel Deschaud}, {and}
  \bibinfo{person}{Fran{\c{c}}ois Goulette}.} \bibinfo{year}{2018}\natexlab{}.
\newblock \showarticletitle{Classification of point cloud scenes with
  multiscale voxel deep network}.
\newblock \bibinfo{journal}{\emph{arXiv preprint arXiv:1804.03583}}
  (\bibinfo{year}{2018}).
\newblock


\bibitem[\protect\citeauthoryear{Sarkar, Hampiholi, Varanasi, and
  Stricker}{Sarkar et~al\mbox{.}}{2018}]%
        {sarkar2018learning}
\bibfield{author}{\bibinfo{person}{Kripasindhu Sarkar},
  \bibinfo{person}{Basavaraj Hampiholi}, \bibinfo{person}{Kiran Varanasi},
  {and} \bibinfo{person}{Didier Stricker}.} \bibinfo{year}{2018}\natexlab{}.
\newblock \showarticletitle{Learning 3d shapes as multi-layered height-maps
  using 2d convolutional networks}. In \bibinfo{booktitle}{\emph{Proceedings of
  the European Conference on Computer Vision (ECCV)}}. \bibinfo{pages}{71--86}.
\newblock


\bibitem[\protect\citeauthoryear{Sedaghat, Zolfaghari, Amiri, and
  Brox}{Sedaghat et~al\mbox{.}}{2016a}]%
        {sedaghat2016orientation}
\bibfield{author}{\bibinfo{person}{Nima Sedaghat},
  \bibinfo{person}{Mohammadreza Zolfaghari}, \bibinfo{person}{Ehsan Amiri},
  {and} \bibinfo{person}{Thomas Brox}.} \bibinfo{year}{2016}\natexlab{a}.
\newblock \showarticletitle{Orientation-boosted voxel nets for 3d object
  recognition}.
\newblock \bibinfo{journal}{\emph{arXiv preprint arXiv:1604.03351}}
  (\bibinfo{year}{2016}).
\newblock


\bibitem[\protect\citeauthoryear{Sedaghat, Zolfaghari, and Brox}{Sedaghat
  et~al\mbox{.}}{2016b}]%
        {DBLP:journals/corr/AlvarZB16}
\bibfield{author}{\bibinfo{person}{Nima Sedaghat},
  \bibinfo{person}{Mohammadreza Zolfaghari}, {and} \bibinfo{person}{Thomas
  Brox}.} \bibinfo{year}{2016}\natexlab{b}.
\newblock \showarticletitle{Orientation-boosted Voxel Nets for 3D Object
  Recognition}.
\newblock \bibinfo{journal}{\emph{CoRR}}  \bibinfo{volume}{abs/1604.03351}
  (\bibinfo{year}{2016}).
\newblock
\showeprint[arxiv]{1604.03351}
\urldef\tempurl%
\url{http://arxiv.org/abs/1604.03351}
\showURL{%
\tempurl}


\bibitem[\protect\citeauthoryear{Shamir}{Shamir}{2008}]%
        {shamir2008survey}
\bibfield{author}{\bibinfo{person}{Ariel Shamir}.}
  \bibinfo{year}{2008}\natexlab{}.
\newblock \showarticletitle{A survey on mesh segmentation techniques}. In
  \bibinfo{booktitle}{\emph{Computer graphics forum}},
  Vol.~\bibinfo{volume}{27}. Wiley Online Library, \bibinfo{pages}{1539--1556}.
\newblock


\bibitem[\protect\citeauthoryear{Shlafman, Tal, and Katz}{Shlafman
  et~al\mbox{.}}{2002}]%
        {shlafman2002metamorphosis}
\bibfield{author}{\bibinfo{person}{Shymon Shlafman}, \bibinfo{person}{Ayellet
  Tal}, {and} \bibinfo{person}{Sagi Katz}.} \bibinfo{year}{2002}\natexlab{}.
\newblock \showarticletitle{Metamorphosis of polyhedral surfaces using
  decomposition}. In \bibinfo{booktitle}{\emph{Computer graphics forum}},
  Vol.~\bibinfo{volume}{21}. Wiley Online Library, \bibinfo{pages}{219--228}.
\newblock


\bibitem[\protect\citeauthoryear{Sinha, Bai, and Ramani}{Sinha
  et~al\mbox{.}}{2016}]%
        {sinha2016deep}
\bibfield{author}{\bibinfo{person}{Ayan Sinha}, \bibinfo{person}{Jing Bai},
  {and} \bibinfo{person}{Karthik Ramani}.} \bibinfo{year}{2016}\natexlab{}.
\newblock \showarticletitle{Deep learning 3D shape surfaces using geometry
  images}. In \bibinfo{booktitle}{\emph{European Conference on Computer
  Vision}}. Springer, \bibinfo{pages}{223--240}.
\newblock


\bibitem[\protect\citeauthoryear{Smith}{Smith}{2017}]%
        {smith2017cyclical}
\bibfield{author}{\bibinfo{person}{Leslie~N Smith}.}
  \bibinfo{year}{2017}\natexlab{}.
\newblock \showarticletitle{Cyclical learning rates for training neural
  networks}. In \bibinfo{booktitle}{\emph{2017 IEEE Winter Conference on
  Applications of Computer Vision (WACV)}}. IEEE, \bibinfo{pages}{464--472}.
\newblock


\bibitem[\protect\citeauthoryear{Su, Maji, Kalogerakis, and Learned-Miller}{Su
  et~al\mbox{.}}{2015}]%
        {su2015multi}
\bibfield{author}{\bibinfo{person}{Hang Su}, \bibinfo{person}{Subhransu Maji},
  \bibinfo{person}{Evangelos Kalogerakis}, {and} \bibinfo{person}{Erik
  Learned-Miller}.} \bibinfo{year}{2015}\natexlab{}.
\newblock \showarticletitle{Multi-view convolutional neural networks for 3d
  shape recognition}. In \bibinfo{booktitle}{\emph{Proceedings of the IEEE
  international conference on computer vision}}. \bibinfo{pages}{945--953}.
\newblock


\bibitem[\protect\citeauthoryear{Sun, Ovsjanikov, and Guibas}{Sun
  et~al\mbox{.}}{2009}]%
        {sun2009concise}
\bibfield{author}{\bibinfo{person}{Jian Sun}, \bibinfo{person}{Maks
  Ovsjanikov}, {and} \bibinfo{person}{Leonidas Guibas}.}
  \bibinfo{year}{2009}\natexlab{}.
\newblock \showarticletitle{A concise and provably informative multi-scale
  signature based on heat diffusion}. In \bibinfo{booktitle}{\emph{Computer
  graphics forum}}, Vol.~\bibinfo{volume}{28}. Wiley Online Library,
  \bibinfo{pages}{1383--1392}.
\newblock


\bibitem[\protect\citeauthoryear{Sun, Page, Paik, Koschan, and Abidi}{Sun
  et~al\mbox{.}}{2002}]%
        {sun2002triangle}
\bibfield{author}{\bibinfo{person}{Yiyong Sun}, \bibinfo{person}{David~Lon
  Page}, \bibinfo{person}{Joon~Ki Paik}, \bibinfo{person}{Andreas Koschan},
  {and} \bibinfo{person}{Mongi~A Abidi}.} \bibinfo{year}{2002}\natexlab{}.
\newblock \showarticletitle{Triangle mesh-based edge detection and its
  application to surface segmentation and adaptive surface smoothing}. In
  \bibinfo{booktitle}{\emph{Proceedings. International Conference on Image
  Processing}}, Vol.~\bibinfo{volume}{3}. IEEE, \bibinfo{pages}{825--828}.
\newblock


\bibitem[\protect\citeauthoryear{Sundar, Silver, Gagvani, and Dickinson}{Sundar
  et~al\mbox{.}}{2003}]%
        {sundar2003skeleton}
\bibfield{author}{\bibinfo{person}{Hari Sundar}, \bibinfo{person}{Deborah
  Silver}, \bibinfo{person}{Nikhil Gagvani}, {and} \bibinfo{person}{Sven
  Dickinson}.} \bibinfo{year}{2003}\natexlab{}.
\newblock \showarticletitle{Skeleton based shape matching and retrieval}. In
  \bibinfo{booktitle}{\emph{2003 Shape Modeling International.}} IEEE,
  \bibinfo{pages}{130--139}.
\newblock


\bibitem[\protect\citeauthoryear{Tam and Lau}{Tam and Lau}{2007}]%
        {tam2007deformable}
\bibfield{author}{\bibinfo{person}{G Tam} {and} \bibinfo{person}{R Lau}.}
  \bibinfo{year}{2007}\natexlab{}.
\newblock \showarticletitle{Deformable model retrieval based on topological and
  geometric signatures.}
\newblock \bibinfo{journal}{\emph{IEEE transactions on visualization and
  computer graphics.}} \bibinfo{volume}{13}, \bibinfo{number}{3}
  (\bibinfo{year}{2007}), \bibinfo{pages}{470--482}.
\newblock


\bibitem[\protect\citeauthoryear{Tchapmi, Choy, Armeni, Gwak, and
  Savarese}{Tchapmi et~al\mbox{.}}{2017}]%
        {tchapmi2017segcloud}
\bibfield{author}{\bibinfo{person}{Lyne Tchapmi}, \bibinfo{person}{Christopher
  Choy}, \bibinfo{person}{Iro Armeni}, \bibinfo{person}{JunYoung Gwak}, {and}
  \bibinfo{person}{Silvio Savarese}.} \bibinfo{year}{2017}\natexlab{}.
\newblock \showarticletitle{Segcloud: Semantic segmentation of 3d point
  clouds}. In \bibinfo{booktitle}{\emph{2017 international conference on 3D
  vision (3DV)}}. IEEE, \bibinfo{pages}{537--547}.
\newblock


\bibitem[\protect\citeauthoryear{Thomas, Qi, Deschaud, Marcotegui, Goulette,
  and Guibas}{Thomas et~al\mbox{.}}{2019}]%
        {thomas2019kpconv}
\bibfield{author}{\bibinfo{person}{Hugues Thomas}, \bibinfo{person}{Charles~R
  Qi}, \bibinfo{person}{Jean-Emmanuel Deschaud}, \bibinfo{person}{Beatriz
  Marcotegui}, \bibinfo{person}{Fran{\c{c}}ois Goulette}, {and}
  \bibinfo{person}{Leonidas~J Guibas}.} \bibinfo{year}{2019}\natexlab{}.
\newblock \showarticletitle{Kpconv: Flexible and deformable convolution for
  point clouds}. In \bibinfo{booktitle}{\emph{Proceedings of the IEEE
  International Conference on Computer Vision}}. \bibinfo{pages}{6411--6420}.
\newblock


\bibitem[\protect\citeauthoryear{Ulyanov, Vedaldi, and Lempitsky}{Ulyanov
  et~al\mbox{.}}{2016}]%
        {ulyanov2016instance}
\bibfield{author}{\bibinfo{person}{Dmitry Ulyanov}, \bibinfo{person}{Andrea
  Vedaldi}, {and} \bibinfo{person}{Victor Lempitsky}.}
  \bibinfo{year}{2016}\natexlab{}.
\newblock \showarticletitle{Instance normalization: The missing ingredient for
  fast stylization}.
\newblock \bibinfo{journal}{\emph{arXiv preprint arXiv:1607.08022}}
  (\bibinfo{year}{2016}).
\newblock


\bibitem[\protect\citeauthoryear{Veli{\v{c}}kovi{\'c}, Cucurull, Casanova,
  Romero, Lio, and Bengio}{Veli{\v{c}}kovi{\'c} et~al\mbox{.}}{2017}]%
        {velivckovic2017graph}
\bibfield{author}{\bibinfo{person}{Petar Veli{\v{c}}kovi{\'c}},
  \bibinfo{person}{Guillem Cucurull}, \bibinfo{person}{Arantxa Casanova},
  \bibinfo{person}{Adriana Romero}, \bibinfo{person}{Pietro Lio}, {and}
  \bibinfo{person}{Yoshua Bengio}.} \bibinfo{year}{2017}\natexlab{}.
\newblock \showarticletitle{Graph attention networks}.
\newblock \bibinfo{journal}{\emph{arXiv preprint arXiv:1710.10903}}
  (\bibinfo{year}{2017}).
\newblock


\bibitem[\protect\citeauthoryear{Verma, Boyer, and Verbeek}{Verma
  et~al\mbox{.}}{2018}]%
        {verma2018feastnet}
\bibfield{author}{\bibinfo{person}{Nitika Verma}, \bibinfo{person}{Edmond
  Boyer}, {and} \bibinfo{person}{Jakob Verbeek}.}
  \bibinfo{year}{2018}\natexlab{}.
\newblock \showarticletitle{Feastnet: Feature-steered graph convolutions for 3d
  shape analysis}. In \bibinfo{booktitle}{\emph{Proceedings of the IEEE
  conference on computer vision and pattern recognition}}.
  \bibinfo{pages}{2598--2606}.
\newblock


\bibitem[\protect\citeauthoryear{Vlasic, Baran, Matusik, and
  Popovi{\'c}}{Vlasic et~al\mbox{.}}{2008}]%
        {vlasic2008articulated}
\bibfield{author}{\bibinfo{person}{Daniel Vlasic}, \bibinfo{person}{Ilya
  Baran}, \bibinfo{person}{Wojciech Matusik}, {and} \bibinfo{person}{Jovan
  Popovi{\'c}}.} \bibinfo{year}{2008}\natexlab{}.
\newblock \showarticletitle{Articulated mesh animation from multi-view
  silhouettes}.
\newblock In \bibinfo{booktitle}{\emph{ACM SIGGRAPH 2008 papers}}.
  \bibinfo{pages}{1--9}.
\newblock


\bibitem[\protect\citeauthoryear{Wang, Cheng, Sohel, Bennamoun, and Li}{Wang
  et~al\mbox{.}}{2019a}]%
        {wang2019normalnet}
\bibfield{author}{\bibinfo{person}{Cheng Wang}, \bibinfo{person}{Ming Cheng},
  \bibinfo{person}{Ferdous Sohel}, \bibinfo{person}{Mohammed Bennamoun}, {and}
  \bibinfo{person}{Jonathan Li}.} \bibinfo{year}{2019}\natexlab{a}.
\newblock \showarticletitle{NormalNet: A voxel-based CNN for 3D object
  classification and retrieval}.
\newblock \bibinfo{journal}{\emph{Neurocomputing}}  \bibinfo{volume}{323}
  (\bibinfo{year}{2019}), \bibinfo{pages}{139--147}.
\newblock


\bibitem[\protect\citeauthoryear{Wang, Pelillo, and Siddiqi}{Wang
  et~al\mbox{.}}{2019c}]%
        {wang2019dominant}
\bibfield{author}{\bibinfo{person}{Chu Wang}, \bibinfo{person}{Marcello
  Pelillo}, {and} \bibinfo{person}{Kaleem Siddiqi}.}
  \bibinfo{year}{2019}\natexlab{c}.
\newblock \showarticletitle{Dominant set clustering and pooling for multi-view
  3d object recognition}.
\newblock \bibinfo{journal}{\emph{arXiv preprint arXiv:1906.01592}}
  (\bibinfo{year}{2019}).
\newblock


\bibitem[\protect\citeauthoryear{Wang, Huang, Hou, Zhang, and Shan}{Wang
  et~al\mbox{.}}{2019b}]%
        {wang2019graph}
\bibfield{author}{\bibinfo{person}{Lei Wang}, \bibinfo{person}{Yuchun Huang},
  \bibinfo{person}{Yaolin Hou}, \bibinfo{person}{Shenman Zhang}, {and}
  \bibinfo{person}{Jie Shan}.} \bibinfo{year}{2019}\natexlab{b}.
\newblock \showarticletitle{Graph attention convolution for point cloud
  semantic segmentation}. In \bibinfo{booktitle}{\emph{Proceedings of the IEEE
  Conference on Computer Vision and Pattern Recognition}}.
  \bibinfo{pages}{10296--10305}.
\newblock


\bibitem[\protect\citeauthoryear{Wang, Asafi, Van~Kaick, Zhang, Cohen-Or, and
  Chen}{Wang et~al\mbox{.}}{2012}]%
        {wang2012active}
\bibfield{author}{\bibinfo{person}{Yunhai Wang}, \bibinfo{person}{Shmulik
  Asafi}, \bibinfo{person}{Oliver Van~Kaick}, \bibinfo{person}{Hao Zhang},
  \bibinfo{person}{Daniel Cohen-Or}, {and} \bibinfo{person}{Baoquan Chen}.}
  \bibinfo{year}{2012}\natexlab{}.
\newblock \showarticletitle{Active co-analysis of a set of shapes}.
\newblock \bibinfo{journal}{\emph{ACM Transactions on Graphics (TOG)}}
  \bibinfo{volume}{31}, \bibinfo{number}{6} (\bibinfo{year}{2012}),
  \bibinfo{pages}{1--10}.
\newblock


\bibitem[\protect\citeauthoryear{Wang, Sun, Liu, Sarma, Bronstein, and
  Solomon}{Wang et~al\mbox{.}}{2019d}]%
        {wang2019dynamic}
\bibfield{author}{\bibinfo{person}{Yue Wang}, \bibinfo{person}{Yongbin Sun},
  \bibinfo{person}{Ziwei Liu}, \bibinfo{person}{Sanjay~E Sarma},
  \bibinfo{person}{Michael~M Bronstein}, {and} \bibinfo{person}{Justin~M
  Solomon}.} \bibinfo{year}{2019}\natexlab{d}.
\newblock \showarticletitle{Dynamic graph cnn for learning on point clouds}.
\newblock \bibinfo{journal}{\emph{ACM Transactions on Graphics (TOG)}}
  \bibinfo{volume}{38}, \bibinfo{number}{5} (\bibinfo{year}{2019}),
  \bibinfo{pages}{1--12}.
\newblock


\bibitem[\protect\citeauthoryear{Williams, Schneider, Silva, Zorin, Bruna, and
  Panozzo}{Williams et~al\mbox{.}}{2019}]%
        {williams2019deep}
\bibfield{author}{\bibinfo{person}{Francis Williams}, \bibinfo{person}{Teseo
  Schneider}, \bibinfo{person}{Claudio Silva}, \bibinfo{person}{Denis Zorin},
  \bibinfo{person}{Joan Bruna}, {and} \bibinfo{person}{Daniele Panozzo}.}
  \bibinfo{year}{2019}\natexlab{}.
\newblock \showarticletitle{Deep geometric prior for surface reconstruction}.
  In \bibinfo{booktitle}{\emph{Proceedings of the IEEE Conference on Computer
  Vision and Pattern Recognition}}. \bibinfo{pages}{10130--10139}.
\newblock


\bibitem[\protect\citeauthoryear{Wu, Song, Khosla, Yu, Zhang, Tang, and
  Xiao}{Wu et~al\mbox{.}}{2015}]%
        {wu20153d}
\bibfield{author}{\bibinfo{person}{Zhirong Wu}, \bibinfo{person}{Shuran Song},
  \bibinfo{person}{Aditya Khosla}, \bibinfo{person}{Fisher Yu},
  \bibinfo{person}{Linguang Zhang}, \bibinfo{person}{Xiaoou Tang}, {and}
  \bibinfo{person}{Jianxiong Xiao}.} \bibinfo{year}{2015}\natexlab{}.
\newblock \showarticletitle{3d shapenets: A deep representation for volumetric
  shapes}. In \bibinfo{booktitle}{\emph{Proceedings of the IEEE conference on
  computer vision and pattern recognition}}. \bibinfo{pages}{1912--1920}.
\newblock


\bibitem[\protect\citeauthoryear{Xu, Zhou, and Qiao}{Xu et~al\mbox{.}}{2019}]%
        {xu2019geometry}
\bibfield{author}{\bibinfo{person}{Mingye Xu}, \bibinfo{person}{Zhipeng Zhou},
  {and} \bibinfo{person}{Yu Qiao}.} \bibinfo{year}{2019}\natexlab{}.
\newblock \showarticletitle{Geometry Sharing Network for 3D Point Cloud
  Classification and Segmentation}.
\newblock \bibinfo{journal}{\emph{arXiv preprint arXiv:1912.10644}}
  (\bibinfo{year}{2019}).
\newblock


\bibitem[\protect\citeauthoryear{Xu, Fan, Xu, Zeng, and Qiao}{Xu
  et~al\mbox{.}}{2018}]%
        {xu2018spidercnn}
\bibfield{author}{\bibinfo{person}{Yifan Xu}, \bibinfo{person}{Tianqi Fan},
  \bibinfo{person}{Mingye Xu}, \bibinfo{person}{Long Zeng}, {and}
  \bibinfo{person}{Yu Qiao}.} \bibinfo{year}{2018}\natexlab{}.
\newblock \showarticletitle{Spidercnn: Deep learning on point sets with
  parameterized convolutional filters}. In
  \bibinfo{booktitle}{\emph{Proceedings of the European Conference on Computer
  Vision (ECCV)}}. \bibinfo{pages}{87--102}.
\newblock


\bibitem[\protect\citeauthoryear{Yang, Litany, Birdal, Sridhar, and
  Guibas}{Yang et~al\mbox{.}}{2020}]%
        {yang2020continuous}
\bibfield{author}{\bibinfo{person}{Zhangsihao Yang}, \bibinfo{person}{Or
  Litany}, \bibinfo{person}{Tolga Birdal}, \bibinfo{person}{Srinath Sridhar},
  {and} \bibinfo{person}{Leonidas Guibas}.} \bibinfo{year}{2020}\natexlab{}.
\newblock \showarticletitle{Continuous Geodesic Convolutions for Learning on 3D
  Shapes}.
\newblock \bibinfo{journal}{\emph{arXiv preprint arXiv:2002.02506}}
  (\bibinfo{year}{2020}).
\newblock


\bibitem[\protect\citeauthoryear{Yavartanoo, Kim, and Lee}{Yavartanoo
  et~al\mbox{.}}{2018}]%
        {DBLP:journals/corr/abs-1811-01571}
\bibfield{author}{\bibinfo{person}{Mohsen Yavartanoo}, \bibinfo{person}{Euyoung
  Kim}, {and} \bibinfo{person}{Kyoung~Mu Lee}.}
  \bibinfo{year}{2018}\natexlab{}.
\newblock \showarticletitle{SPNet: Deep 3D Object Classification and Retrieval
  using Stereographic Projection}.
\newblock \bibinfo{journal}{\emph{CoRR}}  \bibinfo{volume}{abs/1811.01571}
  (\bibinfo{year}{2018}).
\newblock
\showeprint[arxiv]{1811.01571}
\urldef\tempurl%
\url{http://arxiv.org/abs/1811.01571}
\showURL{%
\tempurl}


\bibitem[\protect\citeauthoryear{Zanuttigh and Minto}{Zanuttigh and
  Minto}{2017}]%
        {zanuttigh2017deep}
\bibfield{author}{\bibinfo{person}{Pietro Zanuttigh} {and}
  \bibinfo{person}{Ludovico Minto}.} \bibinfo{year}{2017}\natexlab{}.
\newblock \showarticletitle{Deep learning for 3d shape classification from
  multiple depth maps}. In \bibinfo{booktitle}{\emph{2017 IEEE International
  Conference on Image Processing (ICIP)}}. IEEE, \bibinfo{pages}{3615--3619}.
\newblock


\bibitem[\protect\citeauthoryear{Zhang, Liu, et~al\mbox{.}}{Zhang
  et~al\mbox{.}}{2005}]%
        {zhang2005mesh}
\bibfield{author}{\bibinfo{person}{Hao Zhang}, \bibinfo{person}{Rong Liu},
  {et~al\mbox{.}}} \bibinfo{year}{2005}\natexlab{}.
\newblock \showarticletitle{Mesh segmentation via recursive and visually
  salient spectral cuts}. In \bibinfo{booktitle}{\emph{Proc. of vision,
  modeling, and visualization}}. \bibinfo{pages}{429--436}.
\newblock


\bibitem[\protect\citeauthoryear{Zhi, Liu, Li, and Guo}{Zhi
  et~al\mbox{.}}{2018}]%
        {zhi2018toward}
\bibfield{author}{\bibinfo{person}{Shuaifeng Zhi}, \bibinfo{person}{Yongxiang
  Liu}, \bibinfo{person}{Xiang Li}, {and} \bibinfo{person}{Yulan Guo}.}
  \bibinfo{year}{2018}\natexlab{}.
\newblock \showarticletitle{Toward real-time 3D object recognition: A
  lightweight volumetric CNN framework using multitask learning}.
\newblock \bibinfo{journal}{\emph{Computers \& Graphics}}  \bibinfo{volume}{71}
  (\bibinfo{year}{2018}), \bibinfo{pages}{199--207}.
\newblock


\bibitem[\protect\citeauthoryear{Zhou and Huang}{Zhou and Huang}{2004}]%
        {zhou2004decomposing}
\bibfield{author}{\bibinfo{person}{Yinan Zhou} {and} \bibinfo{person}{Zhiyong
  Huang}.} \bibinfo{year}{2004}\natexlab{}.
\newblock \showarticletitle{Decomposing polygon meshes by means of critical
  points}. In \bibinfo{booktitle}{\emph{10th International Multimedia Modelling
  Conference, 2004. Proceedings.}} IEEE, \bibinfo{pages}{187--195}.
\newblock


\bibitem[\protect\citeauthoryear{Zhu, Chen, Lin, He, Guan, and Zhang}{Zhu
  et~al\mbox{.}}{2019}]%
        {zhu2019random}
\bibfield{author}{\bibinfo{person}{Lei Zhu}, \bibinfo{person}{Weinan Chen},
  \bibinfo{person}{Xubin Lin}, \bibinfo{person}{Li He},
  \bibinfo{person}{Yisheng Guan}, {and} \bibinfo{person}{Hong Zhang}.}
  \bibinfo{year}{2019}\natexlab{}.
\newblock \showarticletitle{Random Walk Network for 3D Point Cloud
  Classification and Segmentation}. In \bibinfo{booktitle}{\emph{2019 IEEE
  International Conference on Robotics and Biomimetics (ROBIO)}}. IEEE,
  \bibinfo{pages}{1921--1926}.
\newblock


\end{thebibliography}


\end{document}